\documentclass[twoside]{article}

\usepackage[accepted]{aistats2024}
\usepackage{booktabs}       
\usepackage{caption}
\usepackage{subcaption}
\usepackage{makecell}
\usepackage{hyperref}
\usepackage{pifont}
\newcommand{\cmark}{\ding{51}}%
\newcommand{\xmark}{\ding{55}}%
\usepackage{comment}
\usepackage[flushleft]{threeparttable}
\usepackage{arydshln}
\usepackage{bbding}
\usepackage{tikz}
\usepackage{comment}
\usepackage{afterpage}
\usepackage{amsmath,amssymb} 
\usepackage{tabto}
\usepackage{xr}
\usepackage{colortbl}
\usepackage{algorithm}
\usepackage{algpseudocode}
\usepackage{newfloat}
\usepackage{listings}
\usepackage{multirow}

\usepackage{array}
\usepackage{longtable}
\usepackage[accsupp]{axessibility}  
\usepackage{upgreek}
\usepackage{tablefootnote}
\usepackage{mathtools}
\usepackage{amsthm}
\usepackage{graphicx}

\usepackage[font=footnotesize]{caption}

\theoremstyle{definition}

\usepackage[round]{natbib}

\begin{document}

%
\runningtitle{XB-MAML: Learning Expandable Basis Parameters for Effective Meta-Learning}

%
\runningauthor{Jae-Jun Lee, Sung Whan Yoon}

\twocolumn[

\aistatstitle{XB-MAML: Learning Expandable Basis Parameters for Effective Meta-Learning with Wide Task Coverage}

\aistatsauthor{ \begin{tabular}{c} Jae-Jun Lee \end{tabular} \And \begin{tabular}{c} Sung Whan Yoon$^\dagger$ \end{tabular} \vspace{0.05in} }

\aistatsaddress{ \begin{tabular}{c} \normalfont\fontfamily{qcr}\small  \{\href{mailto:johnjaejunlee95@unist.ac.kr}{johnjaejunlee95}, \href{mailto:shyoon8@unist.ac.kr}{shyoon8}\}@unist.ac.kr\\ Graduate School of Artificial Intelligence, Ulsan National Institute of Science and Technology (UNIST) \\ \normalfont $^\dagger$ {Corresponding Author}  \end{tabular}}]

\begin{abstract}

Meta-learning, which pursues an effective initialization model, has emerged as a promising approach to handling unseen tasks. However, a limitation remains to be evident when a meta-learner tries to encompass a wide range of task distribution, e.g., learning across distinctive datasets or domains. Recently, a group of works has attempted to employ multiple model initializations to cover widely-ranging tasks, but they are limited in adaptively expanding initializations. We introduce XB-MAML, which learns expandable basis parameters, where they are linearly combined to form an effective initialization to a given task. XB-MAML observes the discrepancy between the vector space spanned by the basis and fine-tuned parameters to decide whether to expand the basis. Our method surpasses the existing works in the multi-domain meta-learning benchmarks and opens up new chances of meta-learning for obtaining the diverse inductive bias that can be combined to stretch toward the effective initialization for diverse unseen tasks.
\end{abstract}

\section{INTRODUCTION} \label{introduction}

Humans have the capability to learn unknown or unseen tasks without explicit prior learning.
When encountering unseen data or learning tasks, humans tap into their meta-knowledge to reason and adapt to the unfamiliar context, drawing upon connections with their past experiences.
This innate adaptability in humans is closely related to the concept of `learning to learn'.
In contrast, deep learning algorithms (\cite{deeplearning}) aim to replicate the humans' learning capability but typically rely on sample-wise likelihood maximization which leans toward memorizing encountered data rather than pursuing meta-knowledge.
As a consequence, modern learning frameworks are prone to fall into critical difficulties, i.e., limited generalization across varying environments.

To tackle the long-lasting challenge, the research field of meta-learning has emerged, drawing inspiration from humans' rapid adaptability to unseen environments by leveraging meta-knowledge across previous experiences.
Specifically, meta-learning aims to effectively solve new tasks by utilizing the commonality acquired from previous learning tasks.
In the domain of classification, meta-learning can be broadly categorized into two main groups: Metric-based methods, encompassing methods such as Matching Network, ProtoNet, TapNet, and TADAM (\cite{matching, protonet, tapnet, tadam}), and optimization-based methods such as Model-Agnostic Meta-Learning (MAML), Reptile, and Meta-SGD (\cite{maml, reptile, meta-sgd}).
Albeit the success of prior works in improving the adaptability to novel tasks, they often fail to handle a wide range of tasks across varying domains, or contexts (\cite{metadatasets}).

By recognizing the hardship of capturing the widely-ranging task distribution through a single meta-trained model parameter,  the concept of employing multiple meta-trained initializations has gained attention in recent times. 
As noticeable trials, TSA-MAML by \cite{tsa-maml} tries to build clusters of similar task parameters to employ per-cluster initialization, and MUSML by \cite{musml} meta-trains multiple subspaces that cover a wider range of task parameters.
The prior works with multi-initializations show obvious limitations in two perspectives: i) The number of initializations is predefined before training, and it is not expandable even if more initializations are required. ii) Utilization of multiple initializations is restricted to selecting one of them so that the combinatorial way of multi-initialization to enlarge the coverage of task distributions is infeasible.

In this paper, we introduce a multi-initialization approach called XB-MAML with two distinctive advantages: i) Expandability of initializations and ii) Combinatorial usage of multiple initializations to provide a better initialization for a given task.
Specifically, XB-MAML incrementally incorporates additional initialization to adaptively cover a wider range of tasks, i.e., the set of initializations is expandable according to the given task distribution.
Also, each initialization works as a `basis' in parameter space, where the meta-trained initializations are linearly combined to form an effective initialization point for the given task.
When the current set of bases falls short of covering task distribution, XB-MAML adaptively employs an additional model parameter, which is likened to increasing the rank of the basis to enable more effective task-specific adaptation across a wide range of complex tasks.
XB-MAML gradually progresses towards the rank of basis that excels in task adaptation and attains performance convergence.
Also, XB-MAML covers the parameter space spanned by the linear combination of the meta-trained bases so that it provides a widened coverage of parameters that cannot be obtained by an individual initialization.
XB-MAML offers a novel strategy to obtain the diverse inductive bias in meta-learning that can be combined to stretch toward the effective initialization for diverse unseen tasks.

In extensive experiments, our XB-MAML shows significant improvements over previous works on the challenging multi-domain few-shot classification in benchmarks datasets: \textit{Meta-Datasets-}ABF/BTAF/CIO.

\section{RELATED WORK} \label{related_works}

\subsection{Recent Advances of Meta-Learning} \label{ssec:recent_ml}

\textbf{Metric-based Method} After the early works of metric-based meta-learning, e.g., Matching Nets by \cite{matching} and ProtoNets by \cite{protonet}, the explicit task-adaptation strategies are adopted in metric learning: TADAM with conditioned representation (\cite{tadam}), TPN with graph-based metric computations (\cite{tpn}), TapNet with task-adaptive projection (\cite{tapnet}). Recently, cross-attention has been shown to be effective in few-shot learning (\cite{can}), and the transformer architecture with in-depth attention draws substantial gains (\cite{spatialformer}).

\textbf{Optimization-based Method} The optimization-based approach aims to train an initialization parameter that is prepared for quick adaptation to the task parameters via bi-level optimization.
Beyond the pioneering work called MAML (\cite{maml}), the computation-efficient variants such as Reptile with the first-order approximation of Hessian (\cite{reptile}) and iMAML with implicit gradients (\cite{imaml}) have been proposed.
As another branch, the probabilistic modeling of the initialization has been explored, e.g., Bayesian MAML by \cite{bmaml} and PLATIPUS by \cite{probabilisticmaml}.
Our XB-MAML is also built on the bi-level optimization by MAML and it incorporates extra initialization by sampling from the Gaussian-based modeling of initializations which is motivated by PLATIPUS.

\subsection{Multi-domain Meta-learning} \label{ssec:multi_domain}

The prior meta-learning works are often limited in training the multi-domain few-shot tasks (\cite{metadatasets}). 
To tackle the issue, MMAML (\cite{mmaml}) employs modulation networks for task grouping, adapting with modulated meta-parameters via gradient updates. HSML (\cite{hsml}) clusters feature-represented tasks hierarchically, adjusting their parameters correspondingly. 
Another approach, ARML (\cite{arml}), utilizes prototype-based relational structures and a meta-knowledge graph to disseminate information within its components. 
These methods primarily focus on task classification or clustering, adapting or selecting tasks within justified groupings to effectively cover a wide array of task distributions.
XB-MAML does not rely on task clustering but meta-trains a set of bases that linearly combine to cover the given task distributions.

\subsection{Multi-initialization Approches} \label{ssec:multi_param}

A group of recent methods has investigated the utilization of multiple initializations to widen the task coverage.
Here, we analyze two distinctive approaches.

\textbf{Clustering of Tasks} \label{sssec:task_similiar}
A method called TSA-MAML (\cite{tsa-maml}) utilizes a pretrained MAML to perform $k$-means clustering to group similar tasks and assign one initialization for each cluster, which is the per-cluster centroid.
Additional episodic learning is then done to further meta-train the centroid initializations.
This process maps each episode to the closest initialization, which can reduce the gap between the initialization and its task-specific model.
However, it raises a crucial concern that TSA-MAML strongly relies on the pretrained MAML which is not tailored for the multi-initialization setting. 
Also, an extensive extra computational burden is required to run the pretraining of MAML.
Lastly, for each task, the best initialization with the smallest loss is selected and meta-updated so that TSA-MAML is limited to naively partitioning the task distribution into clusters.

\begin{figure}[t]
	\captionsetup{justification=centering}
	\centering
	\includegraphics[width=\columnwidth]{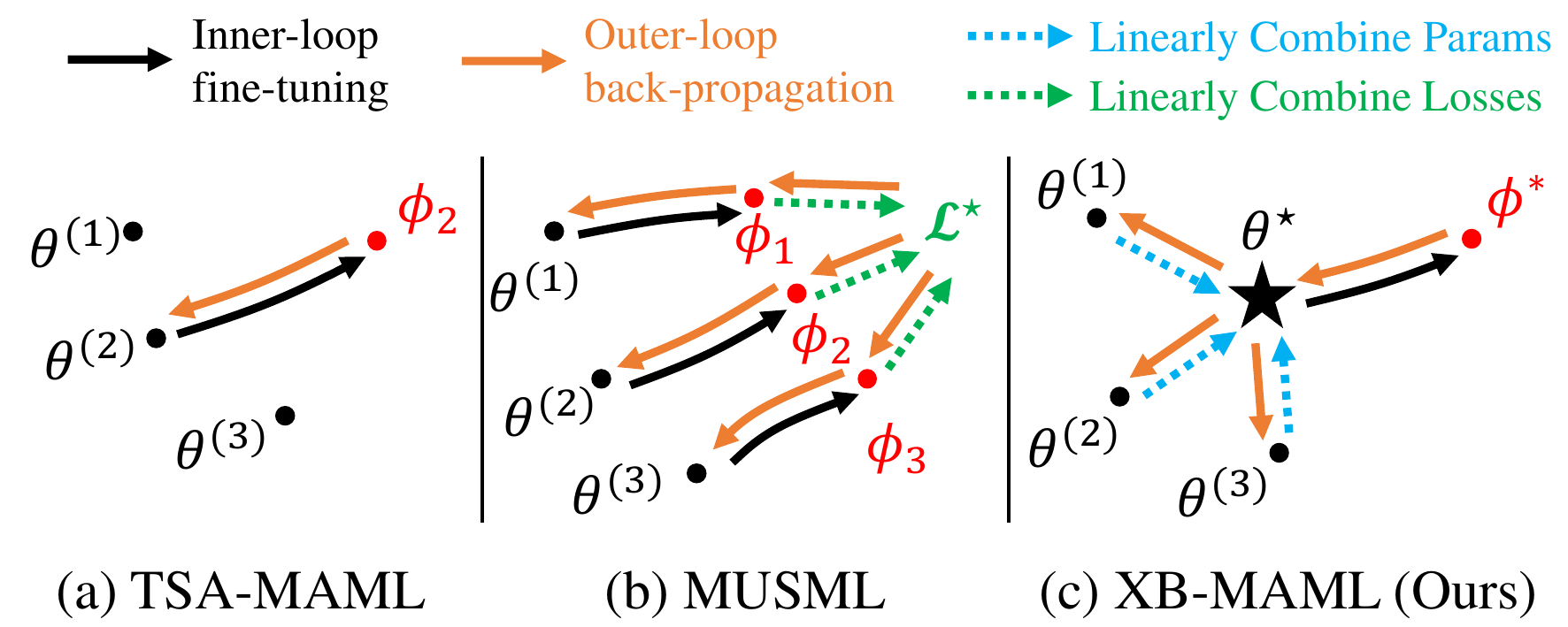}
	\caption{Illustration of bi-level optimization with three initializations $\{\theta^{(m)}\}_{m=1}^{3}$. (a) TSA-MAML selects one initialization with the smallest loss. (b) MUSML separately fine-tunes and meta-updates each initialization. (c) XB-MAML forms the initialization $\theta^{\star}$ via linear combination and jointly meta-updates them.}
	\label{fig:concept}
\end{figure}

\begin{table}[t]
	\centering
	\caption{Comparisons of multi-initialization meta-learners}
	\resizebox{\columnwidth}{!}{
		\begin{tabular}{c|c|c|c}
			\toprule[1.2pt]
			 & \multicolumn{3}{c|}{\textbf{Methods}} \\ 
			 \toprule
			& \textbf{TSA-MAML} & \textbf{MUSML} & \textbf{XB-MAML}\\ 
			\hline
			\textbf{Expandability} & \xmark & \xmark & \cmark \\ 
			\hdashline
			\begin{tabular}[c]{@{}c@{}}\textbf{Use of initial.}\\ \textbf{in training}\end{tabular} & \begin{tabular}[c]{@{}c@{}}Select \& update\\ the best one\end{tabular} & \begin{tabular}[c]{@{}c@{}}Parallelly\\ use \& update\end{tabular} & \begin{tabular}[c]{@{}c@{}}Linear\\ combination\end{tabular} \\ 
			\hdashline
			\begin{tabular}[c]{@{}c@{}}\textbf{Use of initial.}\\ \textbf{in testing}\end{tabular} & \begin{tabular}[c]{@{}c@{}}Use the best one\end{tabular} & \begin{tabular}[c]{@{}c@{}}Use the best one\end{tabular} & \begin{tabular}[c]{@{}c@{}}Linear\\ combination\end{tabular} \\ 
			\hdashline
			\textbf{Extra modules} & \begin{tabular}[c]{@{}c@{}}Pretrained\\ MAML\end{tabular} & \begin{tabular}[c]{@{}c@{}}Subspaces\\ (FC Layers)\end{tabular} & \xmark \\ 
			\specialrule{1.2pt}{2pt}{0pt}
		\end{tabular}}
		\label{tab:comp}
		\vspace{-0.15in}	
\end{table}

\textbf{Learning Subspaces} \label{sssec:subspace}
Another noteworthy approach called MUSML (\cite{musml}) introduces subspace learning and leverages multi-initializations. 
This method quickly learns the significance of subspaces, which consist of simple additional fully connected layers placed right after the feature extractor, effectively serving as a classifier.
MUSML parallelly fine-tunes initializations to the given task and meta-updates them with weighed loss, i.e., a large weight is assigned for the initialization with a smaller meta-loss. 
The process allows multi-initializations to be located properly for covering the task distribution.
As TSA-MAML does, however, MUSML does not address a combinatorial usage of multi-initializations.

It is crucial to acknowledge that the aforementioned methods rely on a predefined number of initializations. It prohibits the prior methods to expand the size of initializations even if additional initializations are required.
In contrast, our XB-MAML adaptively increases initializations if they are needed.
We emphasize that XB-MAML provides a collaborative way of multiple initializations by linearly combining them to build an effective initialization for the given task.
This enables XB-MAML to meta-train the basis of parameters that are required to further stretch out the initialization to the near side of task-specific parameters, which has never been anticipated by the related works.
In Figure \ref{fig:concept} and Table \ref{tab:comp}, the key differences of XB-MAML and the related works are presented.

\section{METHOD: XB-MAML} \label{sec:methods}

\subsection{Preliminaries} \label{ssec:preliminaries}

\textbf{Problem Formulation} \label{sssec:problem}
We follow the episode construction for $N$-way $K$-shot few-shot classification setting with the support/query protocol (\cite{matching}).
Specifically, we sample a batch of tasks $\{\mathcal{T}_i\}_{i=1}^{\mathcal{B}} = \{(\mathcal{S}_{i}, \mathcal{Q}_{i}): i = 1, \cdots , \mathcal{B}\} $, where each task $\mathcal{T}_i$ is sampled from the task distribution $p(\mathcal{T})$. 
Here, $\mathcal{B}$ is the number of tasks in a batch, and $\mathcal{S}_i$ and $\mathcal{Q}_i$ represent the support and query sets of $\mathcal{T}_i$.
To elaborate further, $\mathcal{S}_i$ and $\mathcal{Q}_i$ consist sets of input-label pairs: $\mathcal{S}_i = \{(x^{\mathcal{S}}_{i,j}, y^{\mathcal{S}}_{i,j})\}_{j=1}^{K}$ and $\mathcal{Q}_i = \{(x_{i,j}^{\mathcal{Q}}, y_{i,j}^{\mathcal{Q}})\}_{j=1}^{Q}$, where $\mathcal{S}_i$ contains $K$ samples, referred to as the number of shots, and $Q$ represents the number of query samples.

We introduce additional notations for clarity: $f(\cdot;\theta)$ denotes the output from the model parameterized by $\theta \in\mathbb{R}^{d}$. Also, $\mathcal{L}(\mathcal{D}; \theta) = \frac{1}{|\mathcal{D}|}\sum_{(x, y) \in \mathcal{D}} l(f(x;\theta),y)$ is the averaged loss value for the samples in dataset $\mathcal{D}$ with loss function $l(\cdot, \cdot)$.

\textbf{Review of MAML} \label{sssec:review}
MAML by \cite{maml} involves bi-level optimization of inner and outer loop processes.
Through repetitive learning $\{\mathcal{T}_i\}_{i=1}^{\mathcal{B}}$, MAML optimizes an $\theta$ tailored to the bi-level optimization.
In the inner loop, initialization $\theta$ is rapidly updated for the task using support set $\mathcal{S}_i$. 
We labeled the process as \textbf{Inner-Loop}($\theta, k$), where $k$ is the number of updates in the inner loop (referring Equation \ref{eq:inner_loop}). In the equation, we assume one-step fine-tuning with $k=1$ to obtain task parameter $\phi_i$ from $\theta$.
In the outer loop, the initialization is meta-updated based on the loss incurred by processing query set $\mathcal{Q}_i$ with the fine-tuned parameter (referring Equation \ref{eq:outer_loop}).
\begin{align} 
	&\phi_i  \leftarrow \theta - \alpha \nabla_{\theta} \mathcal{L}(\mathcal{S}_i;\theta) \label{eq:inner_loop} \\
	&\theta^{*} \leftarrow \theta - \frac{\beta}{\mathcal{B}} \nabla_\theta \sum_{i = 1}^{\mathcal{B}}\mathcal{L}(\mathcal{Q}_i; \phi_i),\label{eq:outer_loop}
\end{align}
where $\alpha$ and $\beta$ are the learning rates for inner and outer optimization, respectively.
By repeating the inner and outer loop optimization via task batches, the initialization $\theta$ converges to the meta-trained initialization $\theta^*$ which is ready for quickly adapting to a given novel task.

\subsection{Overview of the Proposed Method} \label{sssec:algorithm}
Our XB-MAML handles multiple initializations so that we newly denote the set of initializations, i.e., $\Theta = \{\theta^{(m)}\}_{m=1}^{{M}}$, where $M$ is the number of initializations.
Here, we describe our method by letting $M$ as a variable, but it does not lose generality in presenting the methodology.
After that, we present the algorithmic way to increase $M$ to incorporate additional initialization.
When focusing on inner and outer optimizations of XB-MAML for task $\mathcal{T}_i$, it starts from computing the loss value by processing the given support samples with each initialization parameter, i.e., $\{\mathcal{L}^{(m)}_i\}_{m=1}^{M}$.
XB-MAML then prepares a new initialization $\mathcal{\theta}_{i}^{\star}$ which is the linear combination of multi-initializations with coefficients $\{\sigma_i^{(m)}\}_{m=1}^{M}$ from softmax computation of the minus losses $\{-\mathcal{L}^{(m)}_i\}_{m=1}^{M}$:
\begin{equation}
	\theta_i^{\star} = \sum_{m=1}^{M}\sigma_i^{(m)} \theta^{(m)}.
\end{equation}
$\theta^{\star}$ is then further fine-tuned within the support set, as done by MAML (referring Equation \ref{eq:inner_loop}), and evaluated on the query sets. 
Along with the loss from queries, we additionally apply a regularization loss $\mathcal{L}_{reg}$ by calculating the dot products between multi-initializations in order to enforce the orthogonality between initializations:
\begin{equation}\label{eq:meta-loss}
	\mathcal{L}_{total, i}^{(m)} = \mathcal{L}(\mathcal{Q}_i; \phi^{\star}_i) + \mathcal{L}_{reg}^{(m)}
\end{equation}
Finally, $\theta^{\star}$ is meta-updated in accordance, where the meta-update chains are further linked to the multi-initializations so that $M$ initializations are eventually meta-updated:
\begin{equation} \label{eq:meta-update}
\theta^{(m)}  \leftarrow \theta^{(m)} -  \frac{\beta}{\mathcal{B}} \sum_{i = 1}^{\mathcal{B}} \nabla_{\theta^{(m)}} \mathcal{L}_{total, i}^{(m)}
\end{equation}
In Equation \ref{eq:meta-loss} and \ref{eq:meta-update}, formulas are given per initialization.
Through the iterative process as done by MAML, XB-MAML meta-trains multiple initializations. 
The pseudocode of the process is described in Algorithm \ref{alg:algorithm1}.
In line 14, Algorithm \ref{alg:algorithm2} determines whether we expand the basis, so let us describe the exact rule for the basis expansion as follows.

\subsection{Expandable Basis Parameters} \label{ssec:XB-MAML}
In this section, we describe how XB-MAML expands the multi-initializations.
For given $\Theta$, each individual initialization can be regarded as linearly independent due to the regularization loss so that it forms a basis in its parameter space. 
According to the formal definition of a basis, $\Theta$ can construct subspace $V \in \mathbb{R}^{M\times d}$ within the parameter space with rank $M$.
XB-MAML aims to utilize the multi-initializations as bases to cover space $V$, which is the key process to obtain a wide range of tasks.
Consequently, when task parameters are hard to be represented on space $V$, we trigger to increase the number of initializations.
It can be understood as the expansion of the rank of $\Theta$ from $M$ $\rightarrow$ $M+1$. 

\begin{algorithm}[t]
    \caption{Training procedures for XB-MAML}
    \textbf{Hyperparameter}: $k$: number of inner loop steps, \\$\mathcal{B}$: batch size, $\gamma$: temperature scaling factor, \\
    $\eta$: regularization hyperparameter\\
    \textbf{Require}: $p(\mathcal{T})$: task distribution, \\
    \textbf{Require}: $\{\mathcal{T}_{i}\}_{i=1}^{\mathcal{B}}$, where $\mathcal{T}_i \sim p(\mathcal{T})$ \\ 
    \textbf{Parameter}: $\Theta = \{\theta^{(m)}\}_{m=1}^{{M}}$: Set of initializations
    
    \begin{algorithmic}[1]
        \State Initialize $\Theta$
        \While{not done}
        	\For{$i=1:\mathcal{B}$}
	        	\State $\{\mathcal{L}^{(m)}\}_{m=1}^{M} = \{\mathcal{L}(\mathcal{S}_i; \theta^{(m)})\}_{m=1}^{M}$
	        	\State $\{\sigma_i^{(m)}\}_{m=1}^M = \frac{\exp(-\mathcal{L}^{(m)} / \gamma)}{\sum_{(m')} \exp(-\mathcal{L}^{(m')} / \gamma)}$
	        	\State $\theta_i^{\star} = \sum_{m=1}^{M}\sigma_i^{(m)} \theta^{(m)} $
	        	\State $\phi_i^{\star}$ = \textbf{Inner-Loop}($\theta^{\star}, k$) 
        	\EndFor
        	\For{$m=1:M$}
	        	\State $\mathcal{L}_{reg}^{(m)} = \frac{\eta}{M-1}\sum_{m!=j} \theta^{(m)} \cdot (\theta^{(j)})^ \intercal$
	        	\State $\{\mathcal{L}_{total, i}^{(m)}\}_{i=1}^{\mathcal{B}} = \{\mathcal{L}(\mathcal{Q}_i; \phi^{\star}_i) + \mathcal{L}_{reg}^{(m)}\}_{i=1}^{\mathcal{B}}$ 
	        	\State $\theta^{(m)}  \leftarrow \theta^{(m)} -  \frac{\beta}{\mathcal{B}} \sum_{i = 1}^{\mathcal{B}} \nabla_{\theta^{(m)}} \mathcal{L}_{total, i}^{(m)}$ 
        	\EndFor
			\State Apply Algorithm \ref{alg:algorithm2}
        \EndWhile
    \end{algorithmic}
    \label{alg:algorithm1}
\end{algorithm}

\begin{algorithm}[tb!]
    \caption{Expanding basis for XB-MAML}
    \textbf{Require}: $\phi^\star_i$: finetuned parameters \\
    \textbf{Require} $I$: index of epoch, $\mathcal{B}$: batch size \\
   	\textbf{Hyperparameter}: $c$: threshold  
    \raggedright
    \begin{algorithmic}[1] 
    	\State Initialize $\mathcal{E}[I] = 0$, count = 0
    	\State Construct Subspace $V = \mathrm{span}\left\{{\theta}^{(1)}, {\theta}^{(2)}, \ldots, {\theta}^{(M)}\right\}$
	    \For{$i=1:\mathcal{B}$}
	    	\State $\phi^\star_{i, proj}$: Projection $\phi^\star_i$ onto subspace $V$
	    	\State $\epsilon = \frac{||\phi^{\star}_i - \phi^{\star}_{i, proj}||^2_2}{||\phi^{\star}_i||_2^2}$
	    	\State $\mathcal{E}[I] += \epsilon / \mathcal{B}$
		\EndFor

		\If{$\mathcal{E}[I] > \mathcal{E}[I-1]$}
			\State count = count + 1 
		\Else 
			\State count = 0
		\EndIf
		\If{count $ > c$}
			\State Add initial model \Comment{Eq.\ref{eq:gaussiansampling}}
		\EndIf
    \end{algorithmic}
    \label{alg:algorithm2}
\end{algorithm}

\begin{figure*}[t]
	\captionsetup{justification=centering}
	\centering
	\includegraphics[width=\textwidth]{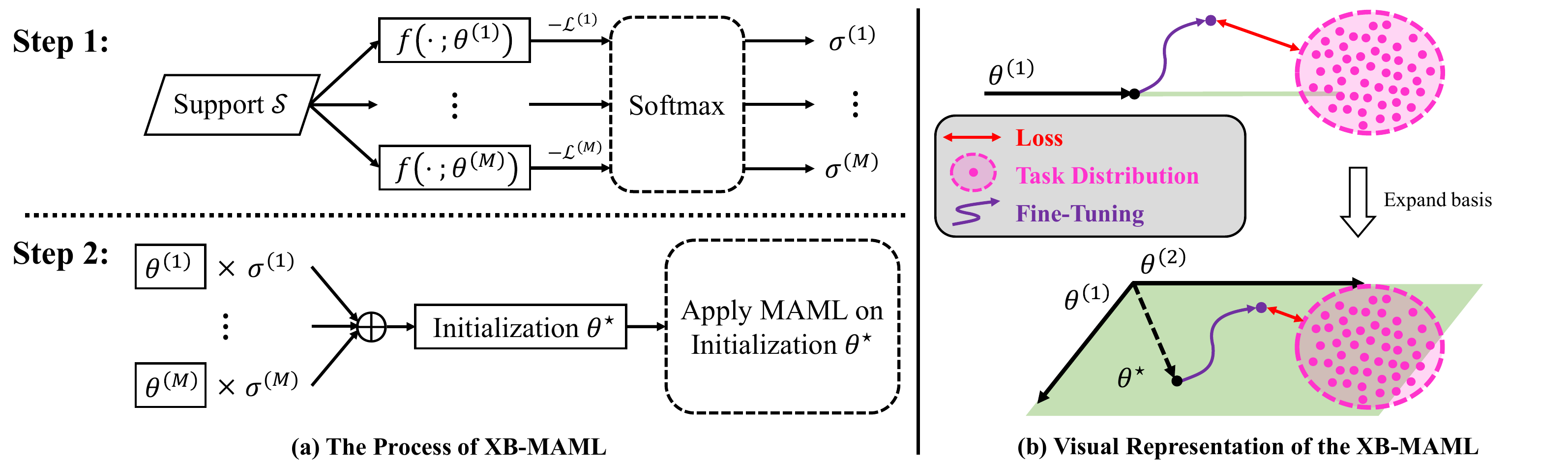}
	\caption{The conceptual illustrations of XB-MAML: \textbf{(a)} outlines the XB-MAML learning process. \\ \textbf{(b)} illustrates the visual representation of the scenario that XB-MAML expands extra basis.}
	\label{fig:algorithm1}
\end{figure*}

\textbf{Condition for Expanding Basis}
We introduced an intuitive metric to determine whether increasing the model parameters is necessary.
As shown in Algorithm \ref{alg:algorithm2}, we first make a subspace $V$ within the set of basis vectors $\{\theta^{(m)}\}_{m=1}^M$.  
After executing the process outlined in lines 3-13 of Algorithm \ref{alg:algorithm1}, it projects the fine-tuned parameters $\phi^{\star}$ into its subspace $S$, which is represented as $\phi^{\star}_{proj}$.
Following this, we calculate the ratio between $||\phi^{\star} - \phi^{\star}_{proj}||^2_2$ and $||\phi^{\star}||^2_2$, denoted as $\epsilon=\frac{||\phi^{\star} - \phi^{\star}_{proj}||^2_2}{ ||\phi^{\star}||_2^2}$, which serves as the primary metric for deciding whether to increase the initial model parameters or not.
When task parameters $\phi^{\star}$ largely deviate from space $V$ spanned by the meta-trained initializations, $\epsilon$ increases.
It says that XB-MAML should increase the rank of basis to cover the diverged task parameters.
After the sufficient training without adding extra initialization, there comes a point where $\epsilon$ starts to rise, signifying that the gap between $\phi^\star$ and $\phi^\star_{proj}$ becomes more significant than the power of $\phi^\star$. 
This phenomenon indicates that the projection error gains prominence.
Based on the intuition, when $\epsilon$ continues to increase during the number of episodes, it becomes necessary to add more initializations.
Specifically, for the current batch at $I$-th epoch, $\epsilon$ values are averaged across the tasks, i.e., $\mathcal{E}[I] += \epsilon / \mathcal{B}$.
When $\mathcal{E}[I]$ is larger than the previous one, i.e., $\mathcal{E}[I]>\mathcal{E}[I-1]$, we increase a counter by +1.
When the counter reaches a certain threshold $c$, it adds extra initialization.
Algorithm \ref{alg:algorithm2} fully describes the condition for basis expansion.

\textbf{Way to Expand Basis}
Instead of naively opting for a random parameter as an additional initialization, we employ a particular strategy involving the sampling of parameters from a probabilistic perspective.
We build a Gaussian distribution with the average of the current initializations for mean, and white noise $\lambda I$, where $\lambda$ controls the variance and $I\in\mathbb{R}^{d\times d}$ is an identity matrix.
We sample additional basis $\theta^{(M+1)}$ from the Gaussian distribution:
\begin{equation} \label{eq:gaussiansampling}
	\theta^{(M+1)} \sim \mathcal{N}(\mu, \lambda I),
\end{equation}
where $\mu$ is the average parameter of $\Theta$.
This probabilistic sampling allows XB-MAML to explore stochastic variants of the new initialization.
Although the newly adopted basis is not forced to be orthogonal to the current bases, we confirm that XB-MAML quickly tunes the new basis to be orthogonal to search additional dimension. 
The overall outline of XB-MAML\footnote{XB-MAML github code is available at \url{https://github.com/johnjaejunlee95/XB-MAML}} is illustrated at Figure \ref{fig:algorithm1}.

\section{EXPERIMENTS} \label{sec:experiments}
In this section, we present a comprehensive overview of experimental settings and results.
We start by providing detailed descriptions of the experimental settings. 
Moreover, our observations demonstrate the improvement of our XB-MAML in few-shot classification for multiple and cross-domain classifications, surpassing the performance of previous studies. 
Additional results, such as single-domain datasets and their cross-domain classification, or experiments on larger backbone, are provided in Appendix \ref{sup_sec:additional_results}.
\begin{table*}[t]
	\caption{5-way 5-shot accuracies on \textit{Meta-Datasets-}ABF with 95$\%$ confidence intervals}
	\label{tab:MetaABF}
	\centering
    \resizebox{0.9\textwidth}{!}{
    \begin{tabular}{c|ccc|c}
		\toprule[1.2pt]
    	Methods & AIRCRAFT & BIRD & FUNGI & Average \\  
    	\midrule
    	MAML (\cite{maml}) & 69.70 $\pm$ 0.33 & 68.36 $\pm$ 0.72 & 53.65 $\pm$ 0.93 & 63.91\\
    	ProtoNet (\cite{protonet}) & 70.07 $\pm$ 0.14 & 71.49 $\pm$ 0.25 & 54.21 $\pm$ 0.31 & 65.26\\
    	HSML (\cite{hsml}) & 68.29 $\pm$ 0.56 & 70.11 $\pm$ 0.85 & 56.28 $\pm$ 1.01 & 64.89 \\
    	ARML (\cite{arml}) & 69.94 $\pm$ 0.78 & 71.55 $\pm$ 0.33 & 53.61 $\pm$ 0.89 & 65.04\\
    	TSA-MAML (5 init) (\cite{tsa-maml}) & 74.67 $\pm$ 0.77 & 71.06 $\pm$ 0.14 & 55.68 $\pm$ 0.27 & 66.14 \\
    	MUSML (3 init) (\cite{musml}) & \textbf{75.46 $\pm$ 0.89} & 70.01 $\pm$ 0.56 & 50.40 $\pm$ 0.75 & 65.29\\
    	\textbf{XB-MAML (4 init)} & {74.39 $\pm$ 0.38} & \textbf{75.17 $\pm$ 0.67} & \textbf{56.85 $\pm$ 0.14} & \textbf{68.80} \\	
    	\midrule
    	MUSML(3 init) + Transduction & \textbf{79.23 $\pm$ 0.98} & 76.21 $\pm$ 0.77 & 58.24 $\pm$ 0.80 & 71.22\\
    	\textbf{XB-MAML(4 init) + Transduction} & {77.62 $\pm$ 0.89} & \textbf{77.78 $\pm$ 0.39} & \textbf{58.34 $\pm$ 0.14} & \textbf{71.24} \\
    	\bottomrule[1.2pt]
    \end{tabular} } \\
    \vspace{.1in}
    
	\caption{5-way 5-shot accuracies on \textit{Meta-Datasets-}BTAF with 95$\%$ confidence intervals}
	\label{tab:MetaBTAF}
	\centering
    \resizebox{0.95\textwidth}{!}{
    \begin{tabular}{c|cccc|c}
    	\toprule[1.2pt]
    	Methods & BIRD & TEXTURE & AIRCRAFT & FUNGI & Average\\  
    	\midrule
    	MAML (\cite{maml})& 67.69 $\pm$ 0.89 & 45.91 $\pm$ 0.54 & 66.93 $\pm$ 0.45 & 50.43 $\pm$ 0.80 & 57.74 \\
    	ProtoNet (\cite{protonet}) & 71.97 $\pm$ 0.74 & 47.65 $\pm$ 0.49 & 69.96 $\pm$ 0.87 & 54.49 $\pm$ 0.41 & 60.02 \\
    	HSML (\cite{hsml}) & 72.01 $\pm$ 0.65 & 49.00 $\pm$ 0.96 & 70.34 $\pm$ 0.68 & 55.21 $\pm$ 0.80 & 61.64 \\
    	ARML (\cite{arml}) & 71.30 $\pm$ 0.44 & 50.48 $\pm$ 0.22 & 70.44 $\pm$ 0.80 & 56.76 $\pm$ 0.56 & 62.25\\
    	TSA-MAML (5 init) (\cite{tsa-maml})& 68.05 $\pm$ 0.94 & 49.61 $\pm$ 0.33 & {73.99 $\pm$ 0.46} & 53.36 $\pm$ 0.20 & 62.25 \\ 
    	MUSML (4 init) (\cite{musml}) & 70.84 $\pm$ 0.32 & 49.63 $\pm$ 0.98 & \textbf{75.73 $\pm$ 0.65} & 49.74 $\pm$ 0.75 & 61.91\\
    	\textbf{XB-MAML (5 init)} & \textbf{75.49 $\pm$ 0.12} & \textbf{50.95 $\pm$ 0.93} & {73.33 $\pm$ 0.16} & \textbf{57.15 $\pm$ 0.71} & \textbf{64.23}\\
    	\midrule
    	MUSML (4 init) + Transduction& 75.32 $\pm$ 0.33 & 52.69 $\pm$ 0.15 & \textbf{77.01 $\pm$ 0.92} & 56.67 $\pm$ 0.81 & 65.45 \\
    	\textbf{XB-MAML (5 init) + Transduction} & \textbf{75.45 $\pm$ 0.90} & \textbf{53.90 $\pm$ 0.85} & 76.11 $\pm$ 0.43 & \textbf{58.34 $\pm$ 0.93} & \textbf{66.20} \\
    	\bottomrule[1.2pt]
    \end{tabular}} \\
    
    \vspace{.1in}    
    \caption{5-way 5-shot accuracies on \textit{Meta-Datasets-}CIO with 95$\%$ confidence intervals}
	\label{tab:MetaCIO}
    \centering
    \resizebox{0.9\textwidth}{!}{
    \begin{tabular}{c|ccc|c}
    	\toprule[1.2pt]
    	Methods & CIFAR-FS & \textit{mini}-ImageNet & Omniglot & Average\\  
    	\midrule
    	MAML (\cite{maml}) & 68.72 $\pm$ 0.43 & 59.84 $\pm$ 0.97 & 96.51 $\pm$ 0.68 & 75.02 \\
    	ProtoNet (\cite{protonet}) & 69.53 $\pm$ 0.72 & 61.40 $\pm$ 0.64 & 97.67 $\pm$ 0.20  & 76.20 \\
    	HSML (\cite{hsml}) & 70.81 $\pm$ 0.97 & 62.45 $\pm$ 0.44 & 96.34 $\pm$ 0.11 & 76.53\\
    	ARML (\cite{arml})  & 70.40 $\pm$ 0.57 & 62.89 $\pm$ 0.48 & 96.80 $\pm$ 0.14 & 76.70\\
    	TSA-MAML (5 init) (\cite{tsa-maml}) & 69.35 $\pm$ 0.26 & 61.20 $\pm$ 0.20 & 98.65 $\pm$ 0.02 & 76.40 \\
    	MUSML (3 init) (\cite{musml}) & 67.97 $\pm$ 0.65 & 59.00 $\pm$ 1.64 & 92.99 $\pm$ 0.41 & 73.41\\
    	\textbf{XB-MAML (6 init)} & \textbf{74.90 $\pm$ 0.35} & \textbf{65.63 $\pm$ 0.12} & \textbf{98.89 $\pm$ 0.09} & \textbf{79.81} \\	
    	\midrule
    	MUSML (3 init) + Transduction & 75.03 $\pm$ 0.38 & 65.54 $\pm$ 0.54 & 96.84 $\pm$ 0.11 & 79.14\\
    	\textbf{XB-MAML (6 init) + Transduction} & \textbf{76.87 $\pm$ 0.73} & \textbf{68.66 $\pm$ 0.53} & \textbf{98.41 $\pm$ 0.04} & \textbf{81.31} \\
    	\bottomrule[1.2pt]
    \end{tabular}  } \\
    
    \raggedright
    \vspace{0.1in}
    
\end{table*}

\subsection{Experimental Settings} \label{ssec:setting}

\textbf{Datasets Description} \label{sssec:hyper_setups1}
We primarily utilized benchmark datasets for our experiments of classification, which have been commonly used in previous works.
Specifically, we employed three multi-domain datasets: \textit{Meta-Datasets-}ABF/BTAF/CIO (\cite{hsml}, \cite{tsa-maml}, \cite{musml}). 
In addition, we also included three experiments on single-domain datasets: CIFAR-FS, \textit{mini}-ImageNet, and \textit{tiered}-ImageNet (\cite{r2d2}, \cite{meta-lstm}, \cite{ssl_meta}). 
Detail descriptions of each dataset is provided in Appendix \ref{supsec:datasets}.

\textbf{Hyperparameters Settings} \label{sssec:hyper_setups2} 
For the learning rates, we use $\alpha=0.05$ and $\beta=0.0007$ for the inner and outer loop.
We conducted training up to 80,000 epochs across multiple datasets, with a batch size of 2.
During meta-validation/test, we evaluated model performance on 600 tasks for each dataset in the multiple datasets. 
Lastly, to ensure a fair comparison, we reproduced all compared approaches via PyTorch, including MAML, ProtoNet, HSML, ARML, TSA-MAML, and MUSML.
Moreover, because some methods used augmentations while others did not, we opted not to apply augmentations except normalization to all methods, including ours. Details of the hyperparameter settings are provided in Appendix \ref{ssec:hyper_settings}.
\vspace{-0.1in}
\subsection{Results} \label{ssec:results}

\textbf{Multiple Domain Datasets Classification} \label{sssec:multi_datasets}
Tables  \ref{tab:MetaABF}, \ref{tab:MetaBTAF}, and \ref{tab:MetaCIO} demonstrate the outstanding results of XB-MAML in the 5-way 5-shot classification. 
This performance gain remains consistent across three multi-domain datasets: \textit{Meta-Datasets-}ABF/BTAF/CIO.
The most significant enhancement is observed in \textit{Meta-Datasets-}CIO, which surpasses previous methods by approximately +3$\%$. 
We also achieve performance gains of around +2$\%$ on remaining datasets.

\textbf{Transduction Setting} \label{sssec:transduction} 
Since the original MUSML (\cite{musml}) applies a transduction setting that utilizes both support and query sets for task-adaptation, we ran experiments in a similar transduction setting for the comparison, as referenced by TPN (\cite{tpn}). TPN modifies loss term with additional regularization by: 
\vspace{-0.1in}
$$\mathcal{L}_{trans} = \mathcal{L}_{total} + \sum_{i=1}^{\mathfrak{N}}\sum_{j=1}^{\mathfrak{N}} W_{(i,j)}||f(x_i; \phi) - f(x_j;\phi)||_2^2$$
where, $W_{(i,j)}=\exp\left(-\frac{1}{2\eta^{2}}\left|\left| f_e(x_i;\phi) - f_e(x_j;\phi)\right|\right|_2^2 \right)$

Here, $\mathfrak{N} =N \times N_s + N_q$, where $N$, $N_s$, and $N_q$ is the number of ways and the number of samples in the support and query sets.
Additionally, $f_e(\cdot;\phi)$ is the output from backbone of the model parameterized by $\phi$, yielding feature representation vectors, while $\eta$ is a scaling factor fixed at a value of $0.25$.
As the result, our method still outperforms MUSML at experiments in transduction settings.

\textbf{Cross-domain Classification} \label{sssec:cross_domain}
We also conducted experiments with cross-domain datasets, where a model is trained on one specific domain and then evaluated on other unseen domains.
The results presented in Table \ref{tab:cross_domain} reveal a substantial performance improvement, with a gain of approximately up to +4$\%$. 
This outcome validates that XB-MAML effectively generates effective initial model parameters in unseen domains.
Here, we abbreviated \textit{Meta-datasets-}ABF/BTAF/CIO as ABF/BTAF/CIO.
\begin{table}[h]
	\caption{5-way 5-shot cross-domain classification}
	\centering
	\resizebox{\columnwidth}{!}{
	\begin{tabular}{c|ccccc}
		\toprule[1.2pt]
		Methods & MAML & TSA-MAML & MUSML & \textbf{XB-MAML}\\  
		\midrule
		ABF$\rightarrow$BTAF & 58.27 & 58.67 & 57.96 & \textbf{61.15} \\
		ABF$\rightarrow$CIO & 60.82 & 62.90 & 62.35 & \textbf{66.73} \\
		\midrule
		BTAF$\rightarrow$ABF & 63.84 & 65.79 & 64.08 & \textbf{68.53} \\
		BTAF$\rightarrow$CIO & 62.69 & 64.24 & 62.99 & \textbf{69.82} \\ 
		\midrule
		CIO$\rightarrow$ABF & 46.38 & 48.58 & 48.55 & \textbf{51.60} \\
		CIO$\rightarrow$BTAF & 44.31 & 46.24 & 45.89 & \textbf{49.03} \\
		\bottomrule[1.2pt]
	\end{tabular}}
	\raggedright
	\centering
	\label{tab:cross_domain}
\end{table}

\section{ANALYSIS} \label{ssec:analysis}

\subsection{t-SNE Visualization} \label{sssec:tsne}

t-SNE plots could be one of the effective choices to illustrate the distributions of model parameters.
We illustrate the t-SNE visualization of the fine-tuned parameters and all initialization parameters across multiple datasets, particularly focusing on the \textit{Meta-Datasets-}ABF experiments.
\begin{figure}[h]
  \begin{subfigure}{0.49\columnwidth} 
    \centering
    \includegraphics[width=\textwidth]{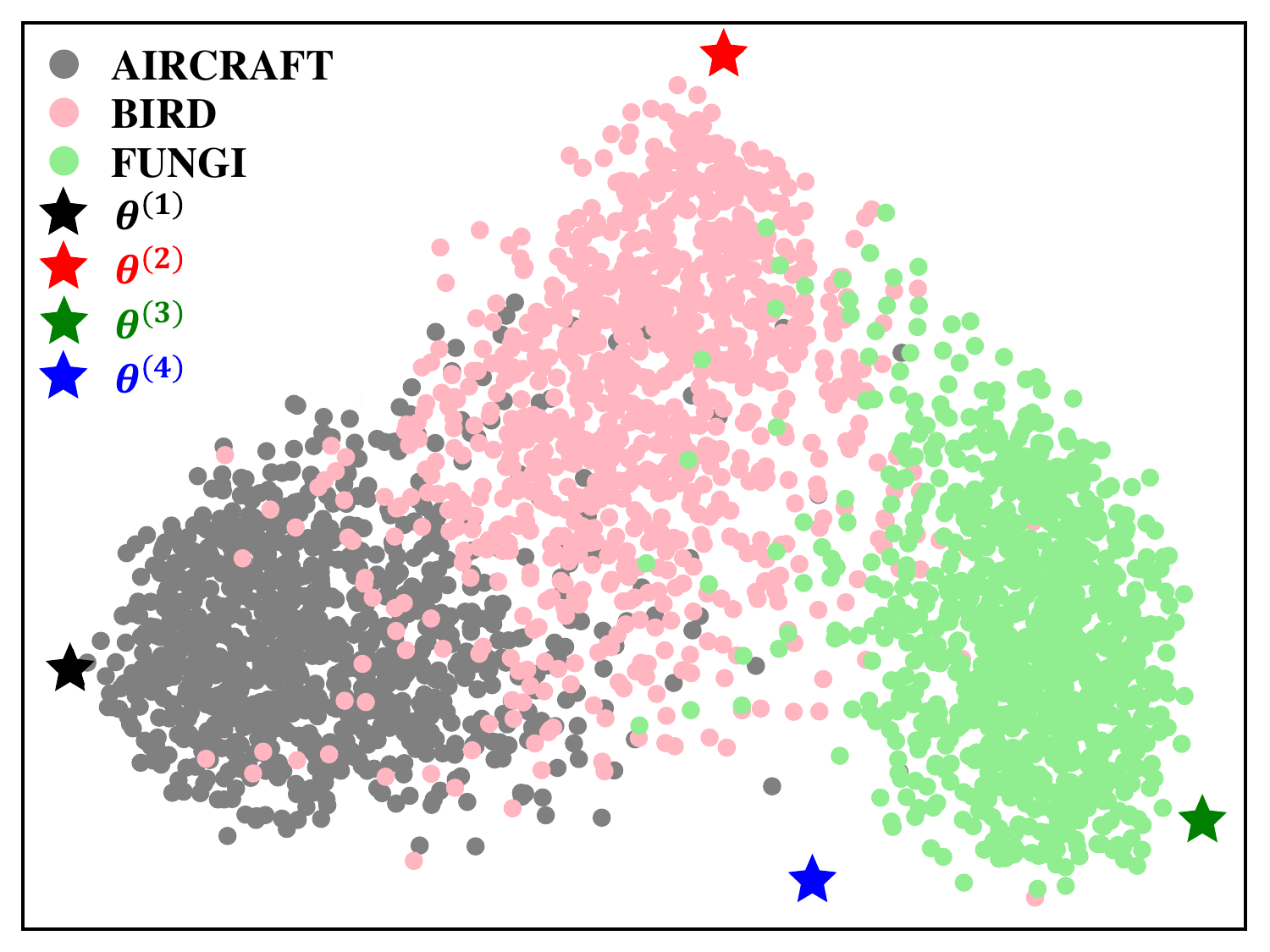}
    \caption{t-SNE Visualization}
    \label{subfig:tsne}
  \end{subfigure}
  \hfill
  \begin{subfigure}{0.49\columnwidth} 
    \centering
    \includegraphics[width=\textwidth]{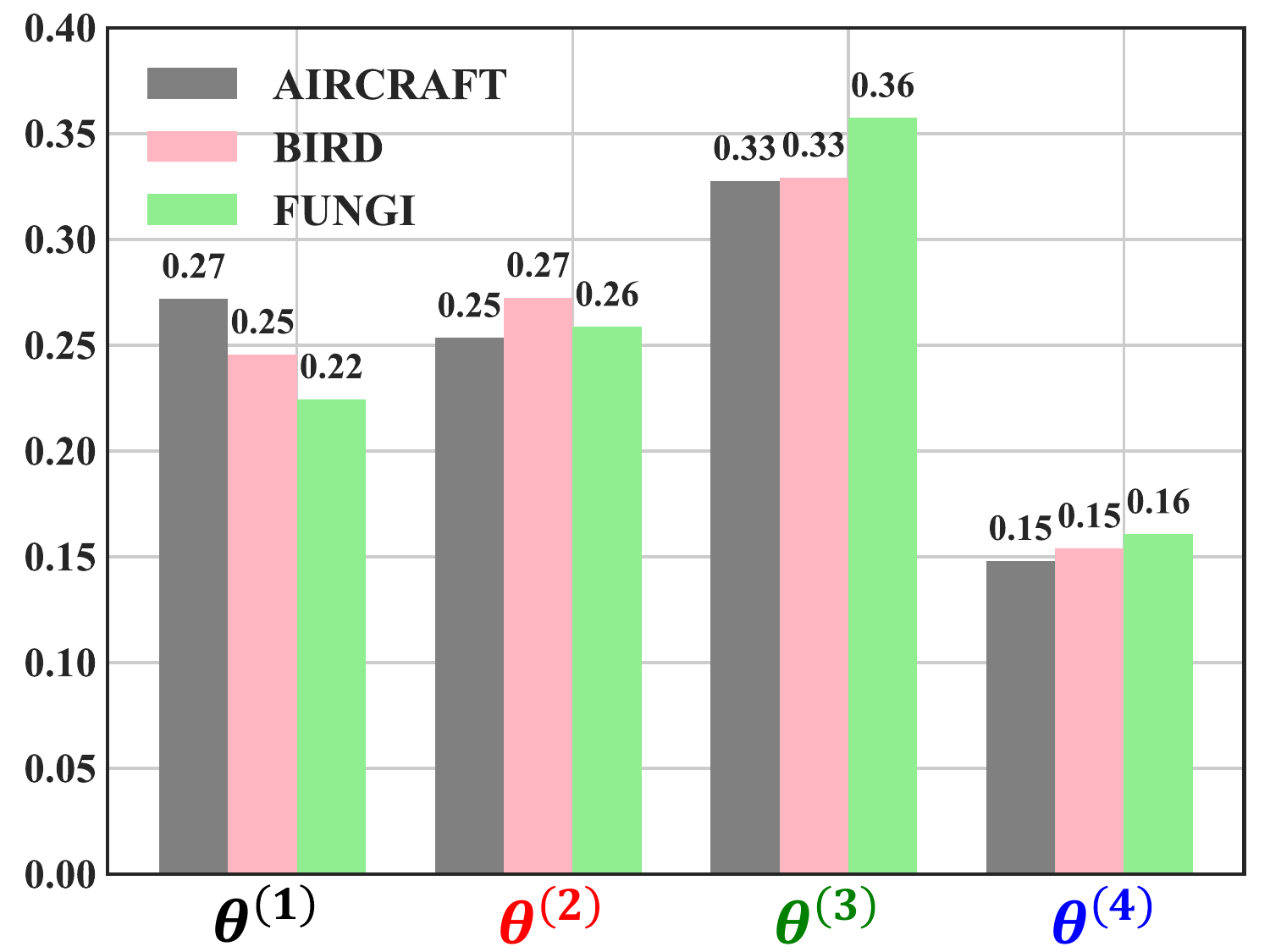}
    \caption{Normalized Coefficients $\sigma$}
    \label{subfig:coeffs}
  \end{subfigure}
  \centering
  \caption{Visualization results in \textit{Meta-Datasets-}ABF. (a) illustrates a t-SNE plot comparing the finetuned parameters and initial model parameters. (b) represents the average normalized coefficients $\sigma$ indicating the impact of each $\theta^{(m)}$ when linearly combined with the datasets.}
  \label{fig:visualization}
\end{figure}
As shown in Figure \ref{fig:visualization}, it is evident that the initialized model parameters are well-distributed (depicted by the colored marker $\star$), while the fine-tuned parameters have been clustered according to their respective datasets (depicted by the colored dots).
Notably, Figure \ref{subfig:coeffs} demonstrates the alignment of coefficients with the t-SNE visualization concept. 
For instance, the coefficient associated with the AIRCRAFT dataset represents the largest value in $\theta^{(1)}$, and this corresponds to the proximity of $\theta^{(1)}$ to the AIRCRAFT cluster.
Furthermore, it is noteworthy that $\theta^{(4)}$ has a comparatively less impact than the others, but still crucial in adapting to various domains, which is positioned in a way that facilitates easy adaptation to any domain.

\subsection{Tendency of $\epsilon$} \label{ssec:metric}
\begin{figure}[h!]
	\captionsetup{justification=centering}
	\centering
	\includegraphics[width=0.9\columnwidth]{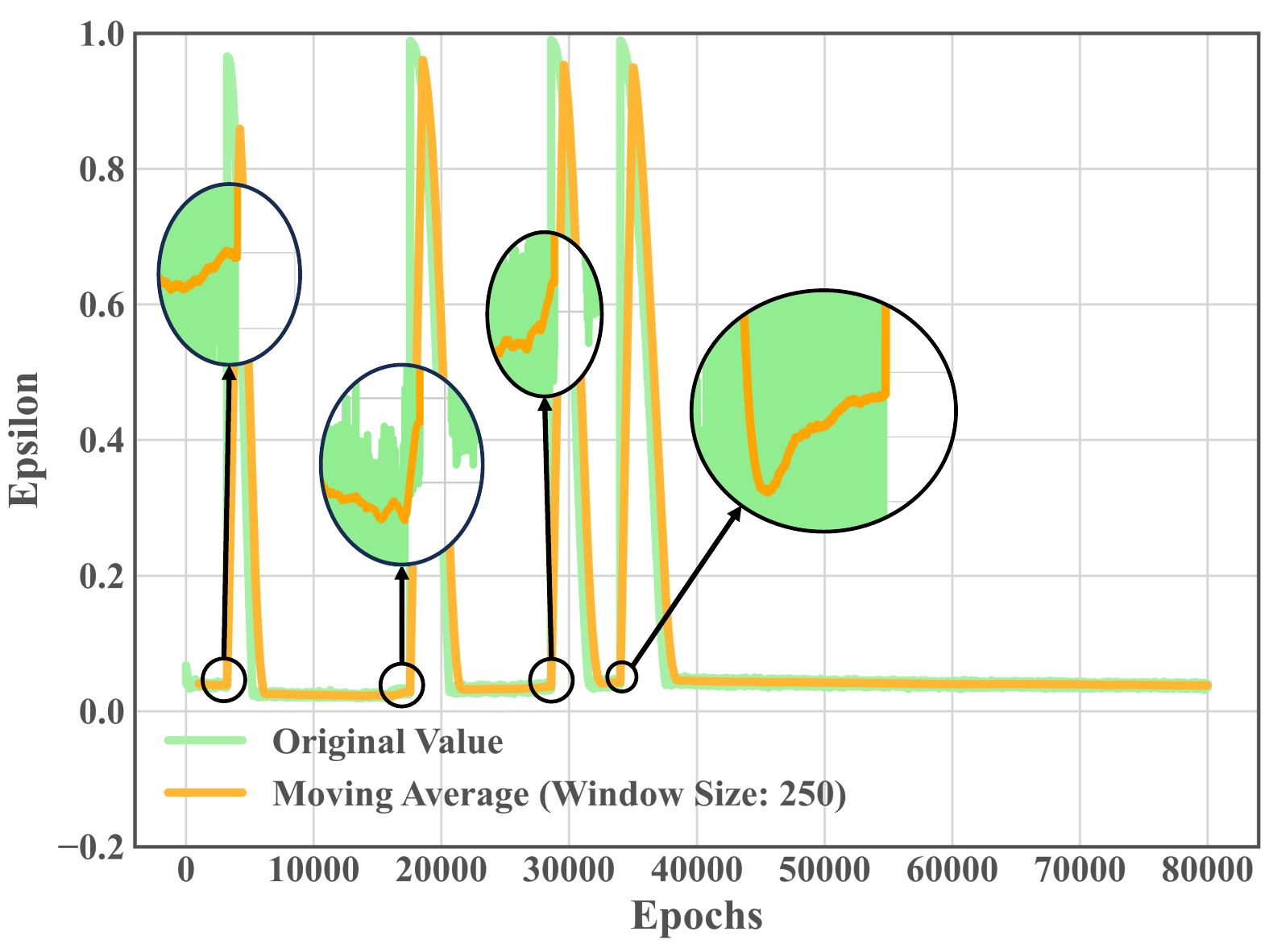}
	\caption{$\epsilon$ on \textit{Meta-Datasets-}BTAF }
	\label{fig:metrics}
\end{figure}

Figure \ref{fig:metrics} displays the impact of our parameter addition metric. When the moving average shows a sustained rise over multiple epochs, we decide to add more initial model parameters. 
We selected a threshold $c$ as 500, to monitor the behavior of $\epsilon$ for a certain steps.
The sudden rise in $\epsilon$, almost up to 1, is due to the instability in constructing the subspaces after adding an initialization. This instability occurs when increasing the number of initializations. 
However, it rapidly decreases and converges within a few steps, indicating that it stabilizes in constructing the basis.
After a few steps of adopting new initializations, $\epsilon$ settles down near to 0, which means that the introduced initializations are sufficient to cover the task distribution.

\subsection{Singular Value Decomposition} \label{sssec:svd}
\begin{figure}[h]
	\captionsetup{justification=centering}
	\centering
	\includegraphics[width=1.00\columnwidth]{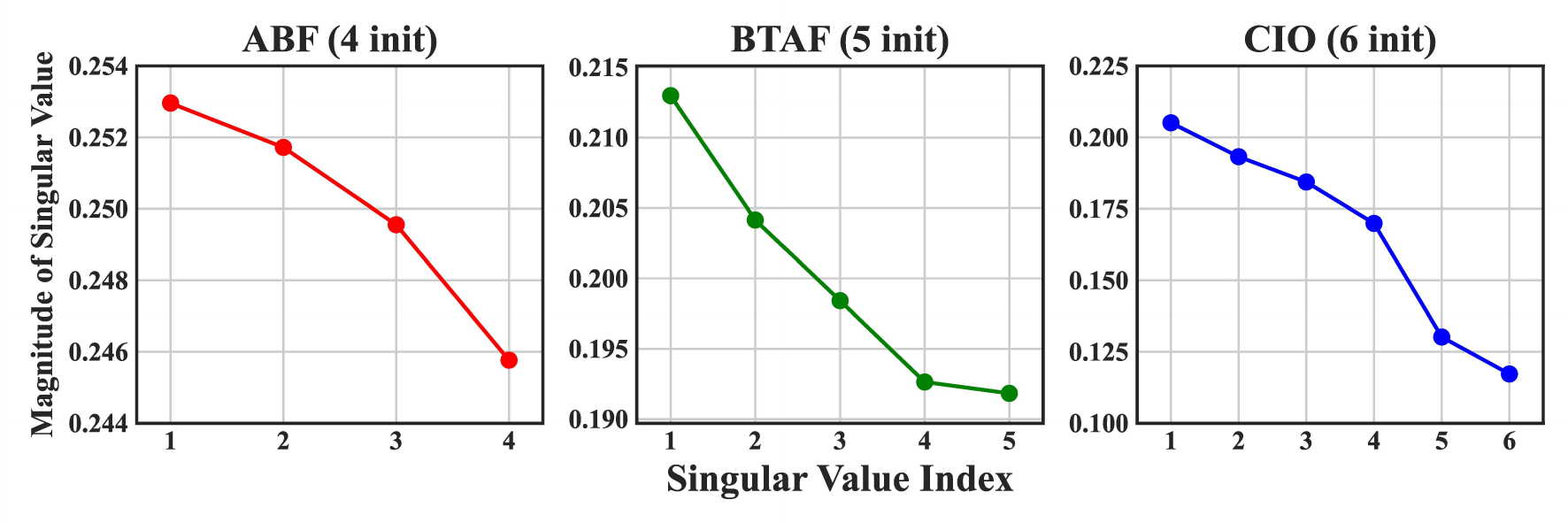}\\
	\caption{Normalized singular values of initializations}
	\label{fig:svd_comparison}
\end{figure}

Figure \ref{fig:svd_comparison} displays the normalized singular values at each datasets from Singular Value Decomposition (SVD) of the multi-initializations.
All singular values fall within the range of approximately $0.1$ to $0.3$. 
These results ensure that no single model parameter dominates, and most model parameters exhibit similarity, in all datasets.
Also, this observation implies that the majority of model parameters has spanned effectively in parameter space.

\subsection{Computational Complexity}
As our method adaptively increases the number of initializations, concerns about computational costs may arise. 
To address these concerns, we conducted a comparative analysis of our approach and two other methods in terms of computational complexity. 
We assume that the computational complexity of the forward and backward processes is $O(n)$, where $n$ represents the number of model weight parameters. 
Additionally, we assume that the backpropagation during the outer loop process is $O(n^2)$, which includes the Hessian matrix computation.
$M$ denotes the number of initializations. 
As a result, our approach demonstrates efficiency, imposing no substantial computational burden in comparison to others, as demonstrated in Table \ref{tab:timecomplexity}.

\begin{table}[h]
	\caption{Analysis of computational complexity}
	\centering
	\resizebox{\columnwidth}{!}{
	\begin{tabular}{c|ccc}
		\toprule[1.2pt]
		Methods & Inner Loop & Outer Loop & Total\\
		\midrule
		TSA-MAML & $O(3n)$ & $O(n^2)$ & $\boldsymbol{O(n^2)}$\\ 
		MUSML & $O(M(n+2n'))^\dagger $& $O(Mn^2)^{\dagger\dagger}$ & $O(Mn^2)$ \\ 
		\textbf{XB-MAML} & $O((M+2)n)$& $O(n^2)$ & $\boldsymbol{O(n^2)}$ \\ 
		\bottomrule[1.2pt]
	\end{tabular}} \\
	\vspace{.1in}
	\raggedright
	\footnotesize{$^\dagger$: $n'$ is the number of subspace weight parameters in MUSML (\cite{musml}}). \\
	\footnotesize{$^{\dagger\dagger}$: Applied relaxation operation, which can be updated model parameters simultaneously (\cite{darts}}).
	\label{tab:timecomplexity}
\end{table}

\subsection{Ablation Studies} \label{ssec:ablation}

\begin{figure}[t] 
	\captionsetup{justification=centering}
    \centering
    \includegraphics[width=\columnwidth]{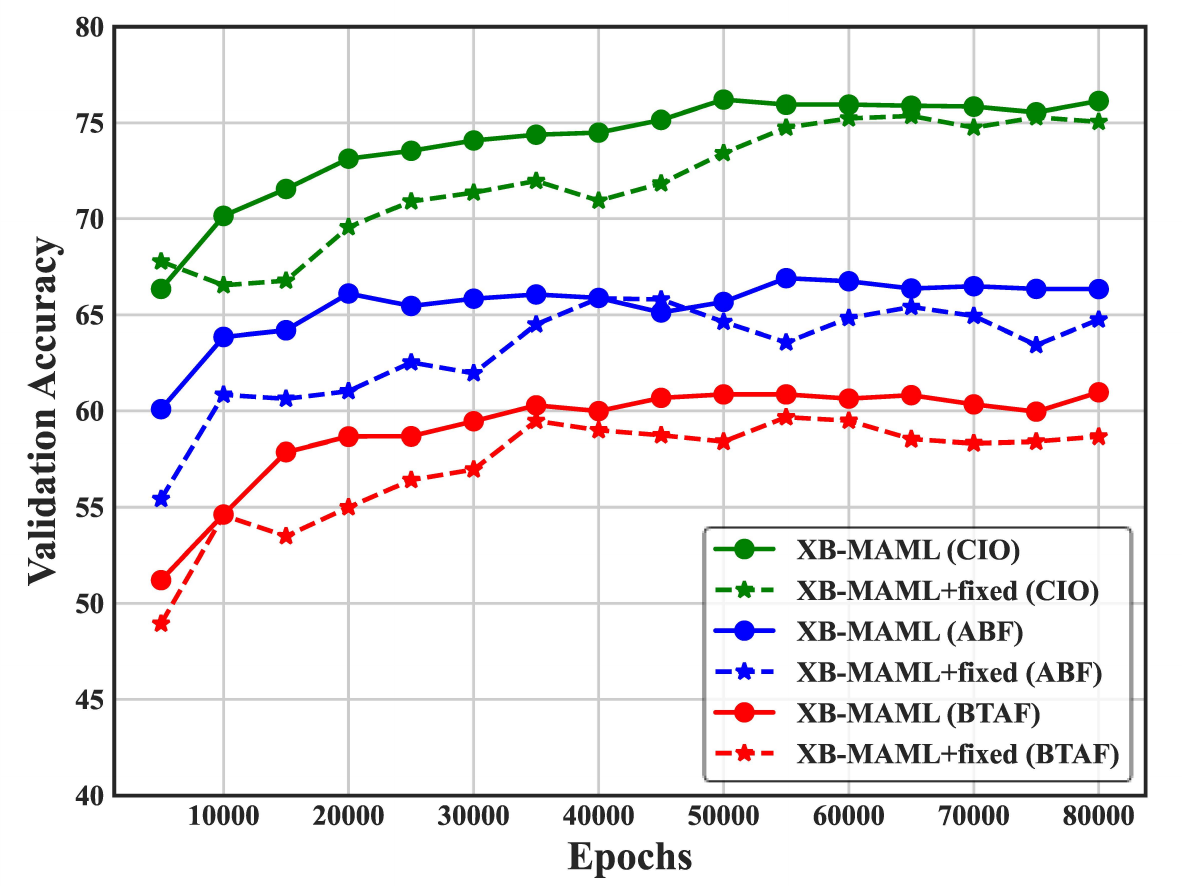}
    \caption{Validation accuracies of XB-MAML and XB-MAML+fixed}
    \label{fig:meta_valid}
\end{figure}

\textbf{Fixed Number of Model Parameters}
In the ablation study, we first introduced a variant of XB-MAML that trains with a fixed number of initialized models, given the same hyperparameter settings. 
For instance, as XB-MAML utilizes 6 initializations in the case of \textit{Meta-Datasets}-CIO, we compare it with a variant of XB-MAML that maintains this number of initializations at 6 from the beginning of the training.
The results presented in Figure \ref{fig:meta_valid} and Table \ref{tab:XB_Comparison} demonstrate that starting with a single initialized model proves to be more efficient than starting with a fixed number of model. 
First, Figure \ref{fig:meta_valid} illustrates this efficiency, which is particularly noticeable in the faster convergence of XB-MAML compared to XB-MAML+fixed.
Also, Table \ref{tab:XB_Comparison} highlights the final outcomes, indicating that XB-MAML outperforms XB-MAML+fixed with the latter representing method that starts with a fixed number of initialized models. \\

\begin{table}[t]
	\centering
	\caption{Results of XB-MAML and XB-MAML+fixed}
	\resizebox{0.95\columnwidth}{!}{
	\begin{tabular}{c|cc}
		\toprule[1.2pt]
		Datasets & \textbf{XB-MAML} & XB-MAML+fixed\\  
		\midrule
		ABF (4 init)& \textbf{68.80 $\pm$ 0.47} & 67.43 $\pm$ 0.38\\ 
		BTAF (5 init)& \textbf{64.23 $\pm$ 0.55} & 62.19 $\pm$ 0.76 \\
		CIO (6 init)& \textbf{79.81 $\pm$ 0.23} & 77.35 $\pm$ 0.14 \\
		\bottomrule[1.2pt]
	\end{tabular}}
	\label{tab:XB_Comparison}
\end{table}

\textbf{Other Choices to Compute $\boldsymbol{\sigma}$}
When answering the reason why we choose `Softmax of minus-loss' $(\exp(-\mathcal{L}))$ method in computing $\sigma$, we opt to use a simple function that outputs a larger positive $\sigma^{(m)}$ when the loss $\mathcal{L}^{(m)}$ is small, ensuring normalized $\sigma$, likewise the attention module. This directly makes us to choose `Softmax of minus-loss'. 
For the ablation study, we additionally compared our method to other choices, including `Equal coefficient', where $\sigma^{(m)}=1/M$, `Inverse loss coefficient', where $\sigma^{(m)}=1/\mathcal{L}^{(m)}$, and `Softmax of minus-loss $\times M$', where $M \exp(-\mathcal{L})$, ensuring that the sum of all values exceeds 1.
As the results in Table \ref{tab:sigma}, our choice is shown to be the best.

\begin{table}[h]
	\centering
	\caption{Results of the various ways to compute $\sigma$}
	\resizebox{\columnwidth}{!}{
		\begin{tabular}{c|>{\centering\arraybackslash}p{1.7cm}|>{\centering\arraybackslash}p{1.7cm}|>{\centering\arraybackslash}p{1.7cm}|>{\centering\arraybackslash}p{2cm}}
            \toprule[1.2pt]
              & $1/M$ & $1/\mathcal{L}$ & $M\exp{(-\mathcal{L})}$ & $\boldsymbol{\exp{(-\mathcal{L}})}$ \\ 
            \midrule
            ABF & 65.32 & 68.30 & 64.67 & \textbf{68.80} \\ 
            BTAF & 61.57 & 63.98 & 61.04 & \textbf{64.23} \\ 
            CIO & 76.18 & 77.34 & 76.62 & \textbf{79.81} \\ 
            \bottomrule[1.2pt]
        \end{tabular}}
	\label{tab:sigma}
\end{table}

\textbf{Sensitivity of $\boldsymbol{c}$} 
As an ablation on $c$, which determines the expansion of bases as described in Algorithm \ref{alg:algorithm2}, we deviated it from 250 to 1,000.
As the results, shown in Table \ref{tab:c_acc}, we found that $c=500$ is the best.
This observation suggests that too small value of $c$ leads to more frequent addition of a new basis, which hinders sufficient training.
Also, too large $c$ could suppress the expansion of the basis, which hampers the ability to cover a wide range of task distribution.

\begin{table}[h]
\centering
        \caption{Accuracies for various choices of $c$}
        \resizebox{\columnwidth}{!}{
        \begin{tabular}{c|ccccc}
            \toprule[1.2pt]
            $c$ & $250$ & $\mathbf{500} $ & $750$ & $1000$ \\ 
            \midrule
            ABF & 67.68 (6 init) & \textbf{68.80 (4 init)} & 67.00 (2 init) & 64.02 (1 init) \\
            BTAF & 63.29 (8 init) & \textbf{64.23 (5 init)} & 62.99 (2 init) & 58.12 (1 init) \\
            CIO & 75.39 (10 init) & \textbf{79.81 (6 init)} & 77.91 (4 init) & 77.46 (3 init)\\
            \bottomrule[1.2pt]
        \end{tabular}}
        \label{tab:c_acc}
\end{table}

\begin{figure}[h]
	\centering
	\includegraphics[width=0.95\columnwidth]{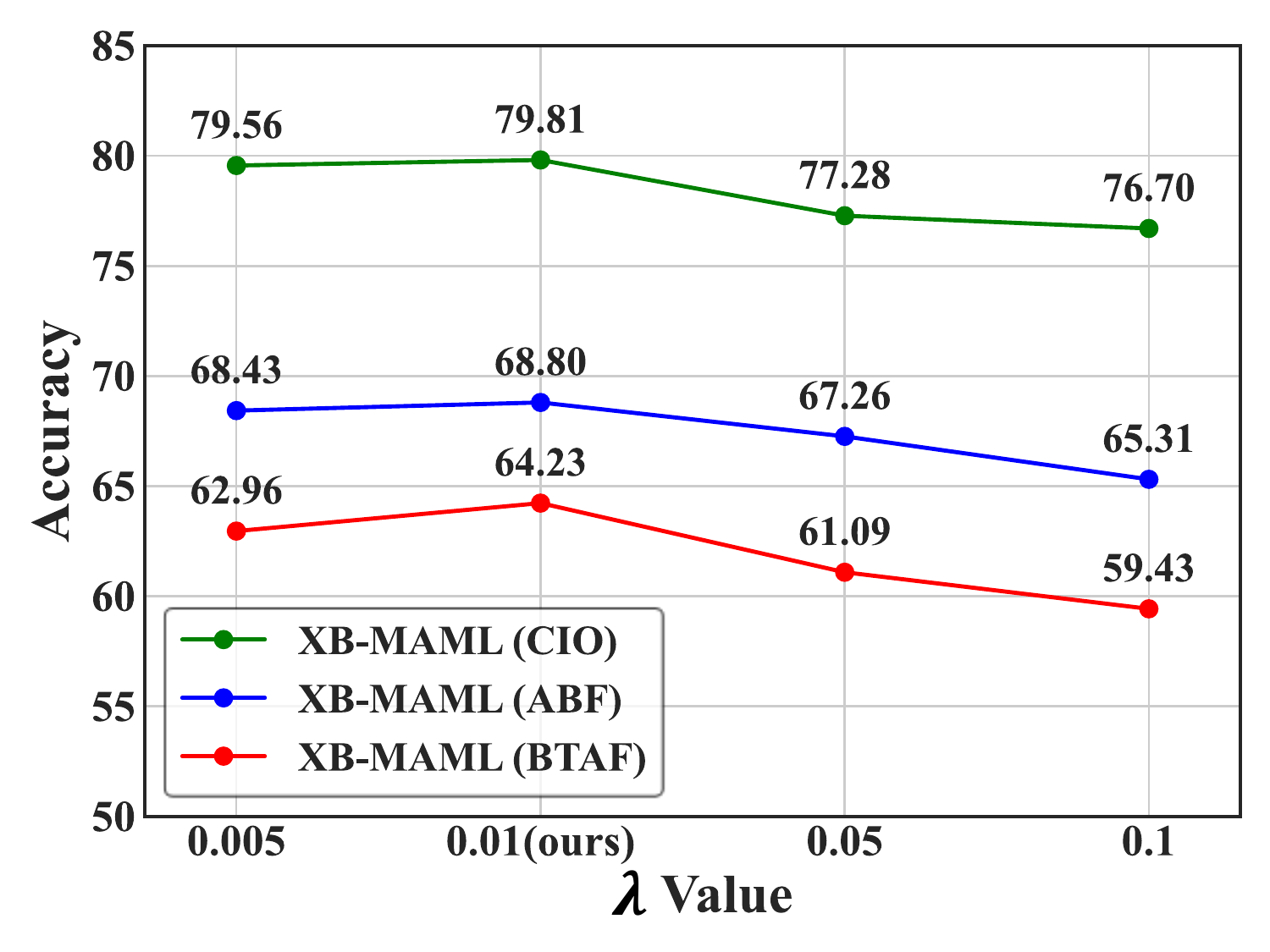}
	\caption{Sensitivity of $\lambda$}
	\label{fig:sen_lam}
\end{figure}

\textbf{Sensitivity of $\lambda$} 
As our method relies on Gaussian sampling when adding additional initializations, the hyperparameter $\lambda$ plays a crucial role in controlling the variance of the resulting Gaussian distribution. 
To explore its impact, we conducted the ablation studies with varying values of $\lambda$, specifically $\lambda = \{0.005, \; 0.01, \; 0.05, \; 0.1 \}$. 
As shown in Figure \ref{fig:sen_lam}, we provided experimental results on multi-domain datasets with several choices of $\lambda$.
These results clearly show that selecting an appropriate value for $\lambda$ is important for achieving better performance. 
When $\lambda$ becomes too large or too small, it adversely impacts the performance.
A large $\lambda$ introduces high uncertainty in the sampling process, hindering effective learning.
Conversely, a small $\lambda$ results in a sampling process that closely samples around the current initializations, which could disrupt the construction of an optimal subspace to cover the task distribution.
We use $\lambda=0.01$, which shows the best performance.

\textbf{The Effect of Number of Initializations}
Here, we analyze the impact of the allowed maximum number of initializations of XB-MAML.
When only 2 initializations are allowed, we stop expanding the initializations beyond 2.
As shown in Table \ref{tab:number_effect}, the results indicate that while performance begins to converge at 4 initializations, further increasing the number of initializations still enhances performance. This suggests that having an adequate number of initialized models is essential.
Also, the results confirm that XB-MAML gradually expands its initializations to reach the optimal rank.

\begin{table}[h]
	\caption{Results with the various maximum number of initializations for \textit{Meta-Datasets-}CIO}
	\centering
	\resizebox{\columnwidth}{!}{
	\begin{tabular}{c|cccccc}
		\toprule[1.2pt]
		Datasets & 1 init& 2 init & 3init & 4 init & 5 init & \textbf{6 init} \\
		\midrule
		CIO & 75.02 & 77.76 & 78.27 & 79.44 & 79.38 & \textbf{79.82} \\ 
		\bottomrule[1.2pt]
	\end{tabular}} \\
	\raggedright
	\label{tab:number_effect}
\end{table}

\textbf{The Effect of $\mathbf{\mathcal{L}_{reg}}$} 
As our approach introduces an additional dot product regularization loss to encourage orthogonality among the initial model parameters, it prompts questions about how this regularization loss influences the span of the set $\Theta$, which acts as a basis, and overall performance. 
Table \ref{tab:without_reg_loss} shows that without $\mathcal{L}_{reg}$, XB-MAML fails to fully span the initializations and to enforce orthogonality (as indicated by cosine similarity), leading to the performance degradation.

\begin{table}[h]
	\caption{Results with $\mathcal{L}_{reg}$ for \textit{Meta-Datasets-}ABF}
	\centering
	\resizebox{\columnwidth}{!}{
	\begin{tabular}{c|cc}
		\toprule[1.2pt]
		Loss & Accuracy  & Cosine Similarity$^\dagger$ \\
		\midrule
		\textbf{With} $\mathbf{\mathcal{L}_{reg}} $& \textbf{68.80} $\mathbf{\pm}$ \textbf{0.47} & $ \textbf{3.95} \mathbf{\times} \mathbf{10^{-5}}$ \\ 
		Without $\mathcal{L}_{reg}$ & 64.32 $\pm$ 0.76 & $7.50 \times 10^{-1}$ \\
		\bottomrule[1.2pt]
	\end{tabular}} \\
	\raggedright
	\vspace{.1in}
	\footnotesize{$^\dagger$: Averaged cosine similarity between initializations}
	\label{tab:without_reg_loss}
\end{table}

\section{CONCLUSION}
We introduce XB-MAML, a novel meta-learning approach that adaptively increases the number of initialized models and refines the initialization points through linear combinations, contributing to more efficient meta-learning. 
The extensive analysis illustrates that XB-MAML competently covers complex and diverse task distributions, particularly in the context of multi-domain and cross-domain classification. Furthermore, we enhanced the performance by treating initialized models as bases, and enforcing orthogonality among them through regularization loss, resulting in the improved performance compared to the absence of such regularization. Finally, our method achieved state-of-the-art results on multi-domain datasets and their cross-domain classifications. We hope this work could provide new perspectives in the research field of meta-learning when solving diverse unseen tasks.

\section*{Acknowledgements}
This work was supported by the Institute of Information \& communications Technology Planning \& Evaluation (IITP) grant funded by the Korea government (MSIT) (No. 2020-0-01336, Artificial Intelligence Graduate School Program (UNIST)), (No. 2021-0-02201, Federated Learning for Privacy Preserving Video Caching Networks) (No. 2023-RS-2022-00156361, Innovative Human Resource Development for Local Intellectualization support program), and the National Research Foundation of Korea (NRF) grant funded by the Korea government (MSIT) (No. 2021R1C1C1012797).

\bibliography{references.bib}

\newpage

\section*{Checklist}
\begin{enumerate}

 \item For all models and algorithms presented, check if you include:
 \begin{enumerate}
   \item A clear description of the mathematical setting, assumptions, algorithm, and/or model. [\textbf{Yes}/No/Not Applicable]
   \item An analysis of the properties and complexity (time, space, sample size) of any algorithm. [\textbf{Yes}/No/Not Applicable]
   \item (Optional) Anonymized source code, with specification of all dependencies, including external libraries. [\textbf{Yes}/No/Not Applicable]
 \end{enumerate}

 \item For any theoretical claim, check if you include:
 \begin{enumerate}
   \item Statements of the full set of assumptions of all theoretical results. [Yes/No/\textbf{Not Applicable}]
   \item Complete proofs of all theoretical results. [Yes/No/\textbf{Not Applicable}]
   \item Clear explanations of any assumptions. [Yes/No/\textbf{Not Applicable}]     
 \end{enumerate}

 \item For all figures and tables that present empirical results, check if you include:
 \begin{enumerate}
   \item The code, data, and instructions needed to reproduce the main experimental results (either in the supplemental material or as a URL). [\textbf{Yes}/No/Not Applicable]
   \item All the training details (e.g., data splits, hyperparameters, how they were chosen). [\textbf{Yes}/No/Not Applicable]
         \item A clear definition of the specific measure or statistics and error bars (e.g., with respect to the random seed after running experiments multiple times). [\textbf{Yes}/No/Not Applicable]
         \item A description of the computing infrastructure used. (e.g., type of GPUs, internal cluster, or cloud provider). [\textbf{Yes}/No/Not Applicable]
 \end{enumerate}

 \item If you are using existing assets (e.g., code, data, models) or curating/releasing new assets, check if you include:
 \begin{enumerate}
   \item Citations of the creator If your work uses existing assets. [\textbf{Yes}/No/Not Applicable]
   \item The license information of the assets, if applicable. [\textbf{Yes}/No/Not Applicable]
   \item New assets either in the supplemental material or as a URL, if applicable. [\textbf{Yes}/No/Not Applicable]
   \item Information about consent from data providers/curators. [\textbf{Yes}/No/Not Applicable]
   \item Discussion of sensible content if applicable, e.g., personally identifiable information or offensive content. [\textbf{Yes}/No/Not Applicable]
 \end{enumerate}

 \item If you used crowdsourcing or conducted research with human subjects, check if you include:
 \begin{enumerate}
   \item The full text of instructions given to participants and screenshots. [Yes/No/\textbf{Not Applicable}]
   \item Descriptions of potential participant risks, with links to Institutional Review Board (IRB) approvals if applicable. [Yes/No/\textbf{Not Applicable}]
   \item The estimated hourly wage paid to participants and the total amount spent on participant compensation. [Yes/No/\textbf{Not Applicable}]
 \end{enumerate}

 \end{enumerate}

\appendix
\onecolumn
\aistatstitle{XB-MAML: Supplementary Materials}

\section{DATASET DESCRIPTION} \label{supsec:datasets}

\subsection{Single-domain Datasets} \label{ssec:single_data}

\textbf{\textit{mini}-ImageNet}:
\textit{mini}-ImageNet, as introduced by \cite{matching}, contains 100 classes with 600 images each, where the images have a size of 84$\times$84$\times$3. 
It is derived from a subset of the ImageNet dataset (\cite{imagenet}). 
The data split follows the protocols proposed by \cite{meta-lstm}, involving 64 classes for the training split, 16 classes for the validation split, and 20 classes for the test split.

\textbf{\textit{tiered}-ImageNet}:
\textit{tiered}-ImageNet, first referenced by \cite{ssl_meta}, is composed of 608 classes distributed across 34 categories, and the images have a size of 84$\times$84$\times$3, which is derived from ImageNet datasets.
Among these categories, 20 are allocated for training, 6 for validation, and 8 for testing. Each category further contains a varying number of classes. In total, the training split contains 351 classes with 448,695 images, the validation split contains 97 classes with 124,261 images, and the test split contains 160 classes with 206,209 images.

\textbf{CIFAR-FS}:
CIFAR-FS is originally comprised by \cite{r2d2}.
This dataset is constructed by randomly sampling from CIFAR-100 datasets (\cite{cifar100}), and it follows a similar data splitting strategy as the \textit{mini}-ImageNet case.
The images in CIFAR-FS have a size of 32$\times$32$\times$3 and are divided into 64/16/20 classes for train/validation/test splits, where each class containing 600 images. 

\subsection{Multiple-domain Datasets} \label{ssec:multi_data}
We present three multi-domain datasets for classification, denoted as \textit{Meta-Datasets-}ABF/BTAF/CIO. 
The dataset labels are defined as follows: `A' for Aircraft, `B' for Bird, `C' for CIFAR-FS, `F' for Fungi, `I' for \textit{mini}-ImageNet, `O' for Omniglot, and `T' for Texture.
All the images in these datasets have been resized to 84$\times$84$\times$3. 
Additionally, as following the work by \cite{hsml}, we have partitioned each of these datasets into 64/12/20 for Train/Validation/Test splits, with the exception of the Omniglot and Texture cases.
We provide descriptions for all datasets except CIFAR-FS and \textit{mini}-ImageNet, which are covered separately in Section \ref{ssec:single_data}.

\textbf{Aircraft}: 
The Aircraft dataset, officially published by \cite{aircraft}, consists of images of 102 different types of aircraft, with 100 images for each type.

\textbf{Bird}:
The Bird dataset, proposed by \cite{bird}, comprises 200 different bird species and a total of 11,788 images. 
In alignment with the protocol presented in the work by \cite{hsml}. We have randomly chosen 100 bird species, each with 60 images. 

\textbf{Fungi}:
The Fungi dataset contains 1,500 distinct fungi species that has more than 100,000 images, which has been introduced by \cite{fungi}.
We filtered out species with fewer than 150 images, following the method described in \cite{hsml}.
Subsequently, we have randomly selected 100 species, each having 150 images.

\textbf{Omniglot}:
The Omniglot dataset proposed by \cite{omniglot} comprises 20 instances within 1,623 characters from 50 different alphabets. Specifically, each instance is handwritten by different persons.
We have splitted 1623 characters into 792/172/659 for Train/Validation/Test.

\textbf{Texture}:
The Texture dataset comprises 5,640 texture images from 47 classes, with each class containing 120 images, which has been provided by \cite{texture}.
In addition, we split these classes into 30/7/10 for Train/Validation/Test.

\vfill

\section{MODEL AND HYPERPARAMETER DESCRIPTION}
\subsection{Model Architecture}
We adopt standard Conv-4 backbone (\cite{matching, maml}) with four convolutional blocks as a feature extractor.
Each block comprises 3$\times$3 convolutional layers with 64 filters, 1 stride, 1 padding, along with a batch normalization layer, ReLU activation functions, and 2$\times$2 max pooling (Conv 3$\times$3 - BatchNorm - ReLU - MaxPool 2$\times$2).
In last, we use a linear classifier that involves adding a fully-connected layer at the end of the Conv-4 backbone. 
All implementations follow the framework provided by Torchmeta (\cite{torchmeta}). All our experiments were run on a single NVIDIA RTX A5000 or A6000 processor. 

\subsection{Hyperparameters Settings} \label{ssec:hyper_settings}
We adopt the following hyperparameter settings described in Table \ref{tab:hyper}.

\begin{table}[h]
	\caption{Hyperparameter Settings}
	\label{tab:hyper}
	\centering
	\resizebox{0.83\textwidth}{!}{
	\begin{tabular}{ccc}
		\toprule[1.2pt]
		Settings & Single-Domain Datasets  & Multi-Domain Datasets \\
		\midrule
		Epochs & 60,000 & 80,000 \\		
		Batch Size (1-shot, 5-shots) & 4, 2 & 4, 2 \\
		Inner-Loop Update Steps (Train, Test) & 3, 7 & 3, 7 \\
		Inner-Loop Learning Rate ($\alpha$) & 0.03 & 0.05 \\
		Outer-Loop Learning Rate ($\beta$)$^*$  & 0.001 & 0.001 \\
		Epoch Threshold ($c$) & 500 & 500 \\ 
		White Noise ($\lambda$) & 0.01 & 0.01 \\
		Temperature Scaling ($\gamma$) & 5 & 8 \\
		Dot Products Regularizer ($\eta$) & 0.0005 & 0.001 \\
		\bottomrule[1.2pt]
	\end{tabular}}\\
	\vspace{.1in}
	\raggedright
	\footnotesize{$^*$: We decrease $\beta$ by multiplying 0.8 at every 20,000 epochs.}
\end{table}

\section{ADDITIONAL RESULTS} \label{sup_sec:additional_results}

\subsection{Single Datasets Classification}
In this section, we present the evaluation results on the single-domain datasets classification, which were excluded from the main paper due to space limitations. 
Table \ref{tab:5w5s_cmt_acc} showcases the 5-way 5-shot performance of XB-MAML in single datasets classification, demonstrating notable improvements compared to other approaches, achieving an improvement of approximately +1$\%$.
These results highlights the effectiveness of XB-MAML in handling single-domain datasets and its outstanding performance in comparison to existing methods, akin to its performance in multi-domain datasets.

\begin{table*}[h]
	\caption{5-way 5-shot Accuracies on single-domain datasets with 95$\%$ confidence intervals}
	\label{tab:5w5s_cmt_acc}
	\centering
	\resizebox{0.9\textwidth}{!}{
	\begin{tabular}{cccc}
		\toprule[1.2pt]
		Methods & CIFAR-FS & \textit{mini}-ImageNet & \textit{tiered}-ImageNet \\
		\midrule
		MAML (\cite{maml}) & 72.41 $\pm$ 0.28 & 63.54 $\pm$ 0.19 & 65.58 $\pm$ 0.15 \\
		ProtoNet (\cite{protonet}) & 72.48 $\pm$ 0.29 & 66.17 $\pm$ 0.15 & 68.32 $\pm$ 0.14 \\
		HSML (\cite{hsml}) & 74.34 $\pm$ 0.37& 65.11 $\pm$ 0.32 & 67.18 $\pm$ 0.45\\
		ARML (\cite{arml}) & 74.76 $\pm$ 0.56 & 65.56 $\pm$ 0.71 & 67.77 $\pm$ 0.63\\
		TSA-MAML (5 init) (\cite{tsa-maml}) & 72.74 $\pm$ 0.19 & 64.29 $\pm$ 0.11 & 66.40 $\pm$ 0.24 \\
		MUSML (2 init) (\cite{musml}) & 73.09 $\pm$ 0.79 & 64.46 $\pm$ 0.32 & 68.05 $\pm$ 0.31 \\
		\textbf{XB-MAML}$^{*}$ & \textbf{75.82 $\pm$ 0.26} & \textbf{66.69 $\pm$ 0.56} & \textbf{68.91 $\pm$ 0.38} \\	
		\bottomrule[1.2pt]
	\end{tabular}} \\
	\raggedright
	\vspace{.1in}
	\footnotesize{$^{*}$: XB-MAML finally adopts 2 for CIFAR-FS, and 4 initializations for both \textit{mini} $\&$ \textit{tiered}-ImageNet.} \\
\end{table*}

\subsection{Additional Results for Cross-domain Classification}
We provide additional results for cross-domain classification, as shown in Table \ref{tab:cd_acc}. Our method exhibits outstanding performance compared to other approaches, with consistent improvements of up to approximately +1$\%$.
It confirms that XB-MAML is consistently effective for the various cases of cross-domain settings.

\begin{table*}[h!]
	\caption{5-way 5-shot accuracies on cross-domain classification}
	\label{tab:cd_acc}
	\centering
	\resizebox{0.9\textwidth}{!}{
	\begin{tabular}{c|>{\centering\arraybackslash}p{1.5cm}|>{\centering\arraybackslash}p{1.5cm}|>{\centering\arraybackslash}p{1.5cm}|>{\centering\arraybackslash}p{1.5cm}|>{\centering\arraybackslash}p{1.5cm}|>{\centering\arraybackslash}p{1.5cm}|>{\centering\arraybackslash}p{1.5cm}|>{\centering\arraybackslash}p{1.5cm}}
		\toprule[1.2pt]
		Train Datasets &\multicolumn{2}{c|}{CIFAR-FS} & \multicolumn{2}{c|}{\textit{mini}-ImageNet} & \multicolumn{2}{c}{\textit{tiered}-ImageNet} \\
		\midrule
		Test Datsets & \textit{mini} & \textit{tiered} & CIFAR & \textit{tiered} & CIFAR & \textit{mini} \\
		\midrule
		MAML (\cite{maml}) & 55.97 & 51.30 & 59.14 & 56.52 & 56.93 & 57.89\\
		TSA-MAML (\cite{tsa-maml}) & 56.43 & 51.87 & 60.21 & 59.34 & 60.34 & 61.84\\
		MUSML (\cite{musml}) &  56.79 & 52.30 & 60.23 & 59.69 & 60.92 & 61.23\\
		\textbf{XB-MAML} &\textbf{57.67}& \textbf{53.40} & \textbf{61.58} & \textbf{61.74} & \textbf{61.91} & \textbf{62.01} \\
		\bottomrule[1.2pt]
	\end{tabular}}
\end{table*}

\subsection{Additional Results in Bigger Backbone}
Given that previous experiments were conducted exclusively using the Conv-4 backbone, there could be some concerns that how XB-MAML would work in a larger backbone. Consequently, we also experimented with ResNet-12, which is also widely utilized as a standard backbone in the meta-learning field. As shown in Table \ref{tab:resnet}, XB-MAML also shows the best performance in the ResNet-12 backbone compared to other methods. Interestingly, the number of bases decreased to 2 or 3. We conjecture that it is due to the sufficient number of parameters already given in the larger backbone that reduces the demands of additional initializations. These results are based on the average of 3 experimental runs.

\begin{table*}[h!]
	\centering
	\caption{Multi-domain classification results with the ResNet-12 backbone} 
	\resizebox{0.5\textwidth}{!}{
		\begin{tabular}{c|ccc}
			\toprule[1.2pt]
			& MAML & TSA-MAML & \textbf{XB-MAML}\\ 
			\midrule
			ABF & 67.76 & 68.28 & \textbf{70.83 (2 init)} \\ 
			BTAF & 63.57 &  65.02 & \textbf{67.60 (2 init)} \\
			CIO & 82.46 & 82.74 & \textbf{84.03 (3 init)} \\
			\bottomrule[1.2pt]
		\end{tabular}
	}
	\label{tab:resnet}
\end{table*}

\section{ADDITIONAL ABLATION STUDIES}
In the recent investigation of the meta-learning research field, the significance of the feature extractor and classifier has become more pronounced, prompting an investigation into which one has a greater impact on performance. 
ANIL by \cite{anil} suggests that rapid adaptation of the classifier can only yield performance results nearly to MAML (\cite{maml}), which encourages the reuse of a single feature representation across tasks.
In contrast, BOIL (\cite{boil}) argues that fine-tuning the feature extractor during the inner loop with fixed classifiers proves exceptionally effective where it leverages the diversity of feature representations.

Hence, we investigate the efficacy of the feature extractor, i.e., body, and classifier, i.e., head, in the context of our approach.
We divide the analysis into two components: increasing the classifier only and increasing the feature extractor only. 
We denote these variants of XB-MAML as XB-MAML-head and XB-MAML-body, where multi-heads are adopted with a single-body initialization and multi-bodies are adopted with a single-head initialization, respectively.
Here, we itemize the related key questions as follows:
\begin{itemize}
	\raggedright
	\item[\textbf{Q1}] Is it essential to introduce a number of classifiers to manage a wide range of task distributions?
	\item[\textbf{Q2}] Is it required to introduce multiple feature extractors to effectively extract the diverse and complex representations across domains?
	\item[\textbf{Q3}] Is there any synergy of multi-body and multi-head in XB-MAML?
\end{itemize}

\begin{table}[h]
	\centering
	\caption{5-way 5-shot accuracies on XB-MAML and its variants with 95$\%$ confidence intervals}
	\label{tab:variants_xb}
	\centering
	\resizebox{\textwidth}{!}{
	\begin{tabular}{c|c|ccc}
		\toprule[1.2pt]
		Domain & Datasets & \textbf{XB-MAML} & XB-MAML-head & XB-MAML-body \\
		\midrule
		\multirow{3}{*}{Single} & CIFAR-FS & \textbf{75.82 $\pm$ 0.26 (2 init)} & 73.17 $\pm$ 0.19 (2 init) & 74.02 $\pm$ 0.46 (3 init)\\
		& \textit{mini}-ImageNet & \textbf{66.69 $\pm$ 0.56 (4 init)} & 63.12 $\pm$ 0.35 (2 init) & 64.12 $\pm$ 0.21 (3 init) \\
		& \textit{tiered}-ImageNet & \textbf{68.91 $\pm$ 0.38 (4 init)} & 67.05 $\pm$ 0.26 (2 init) & 67.54 $\pm$ 0.36 (3 init) \\
		\midrule
		\multirow{3}{*}{\begin{tabular}{c} Multi \\ \end{tabular} } 
		& \textit{Meta-Datasets}-ABF & \textbf{68.80 $\pm$ 0.49 (4 init)} & 66.94 $\pm$ 0.18 (2 init) & 67.84 $\pm$ 0.73 (8 init)\\
		& \textit{Meta-Datasets}-BTAF & \textbf{64.23 $\pm$ 0.27 (5 init)} & 61.24 $\pm$ 0.24 (2 init) & 63.95 $\pm$ 0.49 (8 init)\\
		& \textit{Meta-Datasets}-CIO & \textbf{79.81 $\pm$ 0.11 (6 init)} & 77.50 $\pm$ 0.32 (4 init) & 78.34 $\pm$ 0.13 (10 init) \\
		\bottomrule[1.2pt]
	\end{tabular}}\\
	\raggedright
	\vspace{.1in}
\end{table}

\textbf{Q1} links to the prior understanding of the importance of the classifier part as pointed out by ANIL. On the other hand, \textbf{Q2} is related to the observation by BOIL which argues to diversify the feature extractor part via fine-tuning. Finally, \textbf{Q3} is for clarifying the synergetic effect of multi-body and multi-head via XB-MAML.

Regarding \textbf{Q1}, our findings based on Table \ref{tab:variants_xb} suggest that while incorporating multiple heads, i.e., classifiers, can indeed yield meaningful gains over MAML (referring to the accuracies in the main paper), too many head initializations are not required. 
It is because the classifier primarily comprises fully-connected layers, which are simple linear models that can be efficiently fine-tuned during the inner loop process with a minimal number of initializations. However, it seems that we need a few number of head initializations ranging from 2 to 4.

Conversely, when answering to \textbf{Q2}, the feature extractor requires a greater number of initializations compared to XB-MAML-head, as indicated in Table \ref{tab:variants_xb}. 
Given that the feature extractor is responsible for representing feature distributions, it may require multiple initializations to effectively encompass a broad spectrum of features from distinctive domains. 
This aspect highlights that the feature extractor primarily contributes to the construction of an optimal subspace within the parameter space. 
Moreover, this is also closely tied to the idea that introducing diversity in feature representation is more critical than diversifying selection ability.

Going beyond the previous inquiries, let us think of the answer of \textbf{Q3}. We observe that XB-MAML which meta-trained multi-initializations for both body and head, i.e., an essential aspect of our method, yields the synergetic gains compared to XB-MAML-body and XB-MAML-head.
Moreover, our method reconciles the multi-body and multi-head cases to find fewer initializations than XB-MAML-body but more than XB-MAML-head.
This observation underscores the synergy between increasing the feature extractor and classifier, facilitating a more diverse task adaptation within feature representation and classifier ability.

\end{document}


%

%
\setcounter{section}{0}
\renewcommand{\thesection}{\Alph{section}}
\onecolumn
\aistatstitle{XB-MAML: Supplementary Materials} 

\section{DATASET DESCRIPTION} \label{sec:datasets}

\subsection{single-domain Datasets} \label{ssec:single_data}

\textbf{\textit{mini}-ImageNet}:
\textit{mini}-ImageNet, as introduced by \cite{matching}, contains 100 classes with 600 images each, where the images have a size of 84$\times$84$\times$3. 
It is derived from a subset of the ImageNet dataset \cite{imagenet}. 
Our data split follows the guidelines proposed in the Meta-Learner LSTM by \cite{meta-lstm}, involving 64 classes for the training split, 16 classes for the validation split, and 20 classes for the test split.

\textbf{\textit{tiered}-ImageNet}:
\textit{tiered}-ImageNet, first referenced by \cite{ssl_meta}, is composed of 608 classes distributed across 34 categories, and the images have a size of 84$\times$84$\times$3, which is derived from ImageNet datasets.
Among these categories, 20 are allocated for training, 6 for validation, and 8 for testing. Each category further contains a varying number of classes. In total, the training split contains 351 classes with 448,695 images, the validation split contains 97 classes with 124,261 images, and the test split contains 160 classes with 206,209 images.

\textbf{CIFAR-FS} :
CIFAR-FS is originally comprised by \cite{r2d2}.
This dataset is constructed by randomly sampling from CIFAR-100 datasets \cite{cifar100} and follows similar data splitting strategy as the \textit{mini}-ImageNet.
The images in CIFAR-FS have a size of 32$\times$32$\times$3, and splitted into 64/16/20 classes for train/validation/test datasets, with each class containing 600 images. 

\subsection{Multiple Domain Datasets} \label{ssec:multi_data}
We present three multi-domain datasets for classification, denoted as \textit{Meta-Datasets-}ABF/BTAF/CIO. 
The dataset labels are defined as follows: `A' for Aircraft, `B' for Bird, `C' for CIFAR-FS, `F' for Fungi, `I' for \textit{mini}-ImageNet, `O' for Omniglot, and `T' for Texture.
All the images in these datasets have been resized to 84$\times$84$\times$3. 
Additionally, as following \cite{hsml}, we have partitioned each of these datasets into 64/12/20 for Train/Validation/Test splits, with the exception of the Omniglot and Texture cases.
We provide descriptions for all datasets except CIFAR-FS and \textit{mini}-ImageNet, which are covered separately in Section \ref{ssec:single_data}.

\textbf{Aircraft}: 
The Aircraft dataset, officially published by \cite{aircraft}, consists of images of 102 different types of aircraft, with 100 images for each type.

\textbf{Bird}:
The Bird dataset, proposed by \cite{bird}, comprises 200 different bird species and a total of 11,788 images. 
In alignment with the protocol presented in \cite{hsml}, we have randomly chosen 100 bird species, each with 60 images. 

\textbf{Fungi}:
The Fungi dataset contains 1,500 distinct fungi species that has more than 100,000 images, which has been introduced by \cite{fungi}.
We filtered out species with fewer than 150 images, following the method described in \cite{hsml}.
Subsequently, we have randomly selected 100 species, each having 150 images.

\textbf{Omniglot}:
The Omniglot dataset proposed by \cite{omniglot} comprises 20 instances, within 1,623 characters from 50 different alphabets. Specifically, each instance is handwritten by a different persons.
We splitted 1623 characters into 792/172/659 for Train/Validation/Test, respectively.

\textbf{Texture}:
The Texture dataset comprises 5,640 texture images from 47 classes, with each class containing 120 images, which has been provided by \cite{texture}.
In addition, we split these classes into 30/7/10 for Train/Validation/Test, respectively.

\vfill

\section{MODEL AND HYPERPARAMETER DESCRIPTION}
\subsection{Model Architecture}
We adopt standard Conv-4 backbone (\cite{matching, maml}) with four convolutional blocks as a feature extractor.
Each block comprises 3$\times$3 convolutional layers with 64 filters, 1 stride, 1 padding, along with a batch normalization layer, ReLU activation functions, and 2$\times$2 max pooling. (Conv3$\times$3 - BatchNorm - ReLU - Max2$\times$2)
In last, we use a linear classifier that involves adding a fully-connected layer at the end of the Conv-4 backbone. 
All implementations follow the framework provided by Torchmeta (\cite{torchmeta}). All our experiments were run on a single NVIDIA RTX A5000 or A6000 processor. 

\subsection{Hyperparameters Settings} \label{ssec:hyper_settings}
We adopt the following hyperparameter settings described in Table \ref{tab:hyper}.

\begin{table}[h]
	\caption{Hyperparameters Settings}
	\label{tab:hyper}
	\centering
	\resizebox{0.7\textwidth}{!}{
	\begin{tabular}{ccc}
		\toprule
		Settings & Single-Domain Datasets  & Multi-Domain Datasets \\
		\toprule
		Epochs & 60,000 & 80,000 \\		
		Batch Size (1-shot, 5-shots) & (4, 2) & (4, 2) \\
		Inner-Loop Steps ($k$) (Train, Test) & (3, 7) & (3, 7) \\
		Inner-Loop Learning Rate ($\alpha$) & 0.03 & 0.05 \\
		Outer-Loop Learning Rate ($\beta$)$^\dagger$  & 0.001 & 0.001 \\
		Epoch Threshold ($c$) & 500 & 500 \\ 
		White Noise ($\lambda$) & 0.01 & 0.01 \\
		Temperature Scaling ($\gamma$) & 5 & 8 \\
		Dot Products Regularizer ($\eta$) & 0.0005 & 0.001 \\
		\bottomrule
	\end{tabular}}\\
	\vspace{.1in}
	\raggedright
	\footnotesize{$^\dagger$: We multiply $\beta$ by 0.8 at every 20,000 epochs.}
\end{table}

%
%

\section{ADDITIONAL RESULTS} \label{sup_sec:additional_results}


\subsection{Single Datasets Classification}
In this section, we present the evaluation results on the single-domain datasets, which were excluded from the main paper for brevity. 
Table \ref{tab:5w5s_cmt_acc} showcases the 5ways-5shots performance of XB-MAML in single datasets classification, demonstrating notable improvements compared to other approaches, achieving an improvement of approximately +1$\%$.
These results highlights the effectiveness of XB-MAML in handling single-domain datasets and its outstanding performance in comparison to existing methods, akin to its performance in multi-domain datasets.

%

\begin{table*}[h]
	\caption{5-ways 5-shots Accuracies on single-domain datasets with 95$\%$ confidence intervals}
	\label{tab:5w5s_cmt_acc}
	\centering

	\resizebox{0.9\textwidth}{!}{
	\begin{tabular}{cccc}
		\toprule
		Datasets & CIFAR-FS & \textit{mini}-ImageNet & \textit{tiered}-ImageNet \\
		\midrule
		MAML$^{*}$ \cite{maml} & 72.41 $\pm$ 0.28 & 63.54 $\pm$ 0.19 & 65.58 $\pm$ 0.15 \\
		ProtoNet$^{*}$ \cite{protonet} & 72.48 $\pm$ 0.29 & 66.17 $\pm$ 0.15 & 68.32 $\pm$ 0.14 \\
		HSML$^{*}$ \cite{hsml}& 74.34 $\pm$ 0.37& 65.11 $\pm$ 0.32 & 67.18 $\pm$ 0.45\\
		ARML$^{*}$ \cite{arml}& 74.76 $\pm$ 0.56 & 65.56 $\pm$ 0.71 & 67.77 $\pm$ 0.63\\
		TSA-MAML$^{*}$ (5 init) \cite{tsa-maml} & 72.74 $\pm$ 0.19 & 64.29 $\pm$ 0.11 & 66.40 $\pm$ 0.24 \\
		MUSML$^{*}$ (2 init) \cite{musml}& 73.09 $\pm$ 0.79 & 64.46 $\pm$ 0.32 & 68.05 $\pm$ 0.31 \\
		\textbf{XB-MAML}$^{\dagger}$ & \textbf{75.82 $\pm$ 0.26} & \textbf{66.69 $\pm$ 0.56} & \textbf{68.91 $\pm$ 0.38} \\	
		\bottomrule
	\end{tabular}} \\
	\raggedright
	\vspace{.1in}
	\footnotesize{$^{\dagger}$: CIFAR-FS: 2 init, \textit{mini} $\&$ \textit{tiered}-ImageNet: 4 init} \\
	\footnotesize{$^{*}$: All previous methods are reproduced in PyTorch.} \\
\end{table*}

\subsection{Additional Cross Domain Classification}
We provide additional results for cross domain classification on single-domain, as detailed in Table \ref{tab:cd_acc}. Our method exhibits outstanding performance compared to other approaches, with consistent improvements of up to approximately +1$\%$.

\begin{table*}[h!]
	\caption{5-ways 5-shots accuracies on cross-domain. Here, 95$\%$ confidence intervals are omitted.}
	\label{tab:cd_acc}
	\centering
	\resizebox{0.85\textwidth}{!}{
	\begin{tabular}{c|>{\centering\arraybackslash}p{1.5cm}|>{\centering\arraybackslash}p{1.5cm}|>{\centering\arraybackslash}p{1.5cm}|>{\centering\arraybackslash}p{1.5cm}|>{\centering\arraybackslash}p{1.5cm}|>{\centering\arraybackslash}p{1.5cm}|>{\centering\arraybackslash}p{1.5cm}|>{\centering\arraybackslash}p{1.5cm}}
		\toprule
		Train Datasets &\multicolumn{2}{c|}{CIFAR-FS} & \multicolumn{2}{c|}{\textit{mini}-ImageNet} & \multicolumn{2}{c}{\textit{tiered}-ImageNet} \\
		\midrule
		Test Datsets & \textit{mini} & \textit{tiered} & CIFAR & \textit{tiered} & CIFAR & \textit{mini} \\
		\toprule
		MAML & 55.97 & 51.30 & 59.14 & 56.52 & 56.93 & 57.89\\
		TSA-MAML & 56.43 & 51.87 & 60.21 & 59.34 & 60.34 & 61.84\\
		MUSML &  56.79 & 52.30 & 60.23 & 59.69 & 60.92 & 61.23\\
		\textbf{XB-MAML} &\textbf{57.67}& \textbf{53.40} & \textbf{61.58} & \textbf{61.74} & \textbf{61.91} & \textbf{62.01} \\
		\bottomrule
	\end{tabular}}
\end{table*}


\subsection{Additional Results in Bigger Backbone}
Given that previous experiments were conducted exclusively using the Conv-4 backbone, there could be some concerns that how XB-MAML would function in a larger backbone. Consequently, we also experimented with ResNet-12, which is also widely utilized as a standard backbone in the meta-learning field. As shown in Table \ref{tab:resnet}, XB-MAML also shows the best performance in the ResNet-12 backbone compare to other methods. Since the number of bases dramtically decreased into 2 or 3, it could be due to the sufficient number of parameters already in the model. These results are based on the average of 3 experimental runs.

\begin{table*}[h!]
        \centering
        \caption{Results with ResNet-12}
        \resizebox{0.5\textwidth}{!}{
        \begin{tabular}{c|ccc}
            \toprule
              & MAML & TSA-MAML & XB-MAML\\ 
            \midrule
            ABF  & 67.76 & 68.28 & \textbf{70.83 (2 init)} \\ 
            BTAF & 63.57 &  65.02 & \textbf{67.60 (2 init)} \\
            CIO & 82.46 & 82.74 & \textbf{84.03 (3 init)} \\
            \bottomrule 
        \end{tabular}}
        \label{tab:resnet}
\end{table*}

\section{ADDITIONAL ABLATION STUDY}
In the recent investigation of the meta-learning research field, the significance of the feature extractor and classifier has become more pronounced, prompting an investigation into which one has a greater impact on performance. 
ANIL by (\cite{anil}) suggests that rapid adaptation of the classifier only can yield performance results nearly to MAML (\cite{maml}), which encourages the reuse of a single feature representation.
In contrast, BOIL (\cite{boil}) argues that fine-tuning the feature extractor during the inner loop with fixed classifiers proves exceptionally effective, where it leverages the the change of feature representations.

Hence, we investigate the efficacy of the feature extractor, i.e., body, and classifier, i.e., head, via our approach as an additional ablation study.
We divide the analysis into two components: increasing the classifier only and increasing the feature extractor only. 
We denote these variants of XB-MAML as XB-MAML-head and XB-MAML-body, where multi-heads are adopted with a single-body initialization, and multi-bodies are adopted with a single-head initialization, respectively.
\begin{itemize}
	\raggedright
	\item[\textbf{Q1}] Is it essential to introduce a number of classifiers to manage a wide range of task distributions?
	\item[\textbf{Q2}] Is it required to introduce multiple feature extractors to effectively extract the diverse and complex representations across domains?
	\item[\textbf{Q3}] Is there any synergy of multi-body and multi-head in XB-MAML?
\end{itemize}

\begin{table}[h]
	\centering
	\caption{5-ways 5-shots accuracies on XB-MAML and its variants with 95$\%$ confidence intervals}
	\label{tab:variants_xb}
	\centering
	\resizebox{\textwidth}{!}{
	\begin{tabular}{c|c|ccc}
		\toprule
		Domain & Datasets & \textbf{XB-MAML} & XB-MAML-head & XB-MAML-body \\
		\midrule
		\multirow{3}{*}{Single} & CIFAR-FS & \textbf{75.82 $\pm$ 0.26 (2 init)} & 73.17 $\pm$ 0.19 (2 init) & 74.02 $\pm$ 0.46 (3 init)\\
		& \textit{mini}-ImageNet & \textbf{66.69 $\pm$ 0.56 (4 init)} & 63.12 $\pm$ 0.35 (2 init) & 64.12 $\pm$ 0.21 (3 init) \\
		& \textit{tiered}-ImageNet & \textbf{68.91 $\pm$ 0.38 (4 init)} & 67.05 $\pm$ 0.26 (2 init) & 67.54 $\pm$ 0.36 (3 init) \\
		\midrule
		\multirow{3}{*}{\begin{tabular}{c} Multi \\ \end{tabular} } 
		& ABF & \textbf{68.80 $\pm$ 0.49 (4 init)} & 66.94 $\pm$ 0.18 (2 init) & 67.84 $\pm$ 0.73 (8 init)\\
		& BTAF & \textbf{64.23 $\pm$ 0.27 (5 init)} & 61.24 $\pm$ 0.24 (2 init) & 63.95 $\pm$ 0.49 (8 init)\\
		& CIO & \textbf{79.81 $\pm$ 0.11 (6 init)} & 77.50 $\pm$ 0.32 (4 init) & 78.34 $\pm$ 0.13 (10 init) \\
		\bottomrule

	\end{tabular}}\\
	\raggedright
	\vspace{.1in}
	\footnotesize{* ($n$ init): The Number of initializations based on each method.}
\end{table}

\textbf{Q1} links to the prior understanding of the importance of the classifier part as pointed out by ANIL. On the other hand, \textbf{Q2} is related to the observation by BOIL which argues to diversify the feature extractor part via fine-tuning. Finally, \textbf{Q3} is for clarifying the synergetic effect of multi-body and multi-head via XB-MAML.

Regarding \textbf{Q1}, our findings based on the Table \ref{tab:variants_xb} suggest that while incorporating multiple heads, i.e., classifiers, can indeed yield meaningful gains over MAML (referring to the accuracies in the main paper), too many head initializations are not required. 
It is because the classifier primarily comprises fully-connected layers, which are simple linear models that can be efficiently fine-tuned during the inner loop process with a minimal number of initializations. However, it seems that we need a few number of head initializations ranging from 2 to 4.

Conversely, considering a potential answer to \textbf{Q2}, the feature extractor requires a greater number of initializations compared to XB-MAML-head, as indicated in Table \ref{tab:variants_xb}. 
Given that the feature extractor is responsible for representing feature distributions, it may require multiple initializations to effectively encompass a broad spectrum of features from distinctive domains. 
This aspect highlights that the feature extractor primarily contributes to the construction of an optimal subspace within the parameter space. 
Moreover, this is also closely tied to the idea that introducing diversity in feature representation is more critical than diversifying selection ability.

Going beyond the previous inquiries, let us think of the answer of \textbf{Q3}. We observe that XB-MAML which meta-trained multi-initializations for both body and head, i.e., an essential aspect of our method, yields the synergetic gains compared to XB-MAML-body and XB-MAML-head.
Moreover, our method reconciles the multi-body and multi-head cases to find fewer initializations than XB-MAML-body but more than XB-MAML-head.
This observation underscores the synergy between increasing the feature extractor and classifier, facilitating a more diverse task adaptation within feature representation and classifier ability.

\newpage
